\documentclass[preprint]{elsarticle}

\usepackage{lineno,hyperref}
\usepackage{graphicx}
\usepackage{paralist}
\usepackage{booktabs} 
\usepackage{multirow}
\usepackage{multicol}
\usepackage{arydshln}
\usepackage{subfigure}
\usepackage{xcolor}

\newcommand{\emotion}[1]{\texttt{#1}}

\modulolinenumbers[5]
\journal{Journal of Information Processing and Management}

\newcommand{\partitle}[1]{\vspace{1mm}\noindent\textbf{#1}}

\biboptions{authoryear}

\begin{document}

\begin{frontmatter}


\title{Mental Disorders on Online Social Media \\ Through the Lens of Language and Behaviour: \\ Analysis and Visualisation}

\author{Esteban A. R\'issola}
\address{Faculty of Informatics, Universit\`a della Svizzera italiana, Lugano, Switzerland}

\author{Mohammad Aliannejadi}
\address{IRLab, University of Amsterdam, Amsterdam, The Netherlands}

\author{Fabio Crestani}
\address{Faculty of Informatics, Universit\`a della Svizzera italiana, Lugano, Switzerland}





\begin{abstract}
\par Due to the worldwide accessibility to the Internet along with the continuous advances in mobile technologies, physical and digital worlds have become completely blended, and the proliferation of social media platforms has taken a leading role over this evolution.
In this paper, we undertake a thorough analysis towards better visualising and understanding the factors that characterise and differentiate social media users affected by mental disorders. We perform different experiments studying multiple dimensions of language, including vocabulary uniqueness, word usage, linguistic style, psychometric attributes, emotions' co-occurrence patterns, and online behavioural traits, including social engagement and posting trends. 
\par Our findings reveal significant differences on the use of function words, such as adverbs and verb tense, and topic-specific vocabulary, such as biological processes. As for emotional expression, we observe that affected users tend to share emotions more regularly than control individuals on average. Overall, the monthly posting variance of the affected groups is higher than the control groups. Moreover, we found evidence suggesting that language use on micro-blogging platforms is less distinguishable for users who have a mental disorder than other less restrictive platforms. In particular, we observe on Twitter less quantifiable differences between affected and control groups compared to Reddit.

\end{abstract}


\end{frontmatter}


\section{Introduction}
\label{sec:introduction}
Over the past decade, there has been increasing interest in modelling and detecting mental state changes by analysing online digital footprints~\citep{Correia:2020}. Among the most relevant reasons for this there is the fact that public health systems are limited in their abilities to deal with the large volume of cases they face daily. Yet online social media platforms are spreading and thus, changing the dynamics of mental health assessment~\citep{Skaik:2020, Chancellor:2020,Thieme:2020}. 
On a regular basis, individuals use social media platforms to share their interests~\citep{Zarrinkalam:2018}, personal daily life experiences~\citep{Khodabakhsh:2018,Saha:2021}, and even their moods and feelings~\citep{Prieto:2014}. Using such user-generated content, the development of low-cost, unobtrusive prognostic technologies to assess risks and aid at making decisions has excellent potential to allow healthcare practitioners to perform preliminary screenings and raise awareness of mental health outcomes at a large scale~\citep{Ramirez_Cifuentes:2020,Yates:2017}. 


\par Researchers studying the connection between psychology and language have found many cues about individuals' mental states (including their social and emotional condition), which can be assessed by examining their linguistic expression patterns. In this regard, language use could act as an indicator of the current mental state~\citep{Coppersmith:2015b,Rissola:2019a}, personality~\citep{Neuman:2016,Rissola:2019b} and even personal values~\citep{Boyd:2015,Aliannejadi:2020} of individuals. According to \cite{Pennebaker:2003}, the main reason behind this is that latent mental parameters are encoded in the words that individuals employ to express themselves. Moreover, several studies have shown that individuals' online behavioural traits, such as social engagement, collectively provide significant cues about their mental state. On many occasions, such traits can be considered as proxies, although imperfect, for findings in the medical literature related to positive mental health concerns outcomes~\citep{Coppersmith:2014a,Choudhury:2013a,Reece:2017}. 
For instance, literature on depression indicates that individuals presenting signs of depression exhibit significant changes in their online activity levels in the evening and before dawn~\cite{Thij:2020}.


\par Constraints dictated in real-life settings, such as cost and time, hinder the possibility to achieve a timely and effective personal diagnosis. Recently, initiatives such as the Strategic Workshop on Information Retrieval in Lorne~\citep{Culpepper:2018} (SWIRL) spotlight the possibility of cross-disciplinary collaborations with several scientific disciplines, including psychology. In particular, they highlight that the core principles of Information Retrieval, and related fields, have been gradually extended and applied in domains that a few years back were not easy to conceive or imagine. In this regard, the Computational Linguistics and Clinical Psychology (CLPsych)~\citep{Coppersmith:2015a} and the Early Risk Prediction on the Internet (eRisk)~\citep{Losada:2018,Losada:2019} experimental frameworks have fostered collective efforts worldwide to leverage social media data to develop models for estimating the occurrence of signs related to mental disorders.

\par These initial efforts to address the automatic identification of potential cases of mental disorders on social media have mainly modelled the problem as classification~\citep{Masood:2019,Rissola:2021}. Researchers participating in these workshops have examined a wide variety of methods to identify positive cases\footnote{Throughout the article, we use the term \textit{positive} to refer to individuals who are potentially suffering from mental health concerns such as depression, anorexia, self-harm or post-traumatic stress disorder (PTSD). In contrast, individuals who are in the \textit{control} group do not suffer from any of the previously cited disorders.}~\citep{Burdisso:2019,Cacheda:2019}; however, not much insight has been given as to why a system succeeds or fails. With such insights, the models and features used in those studies could be analysed and motivated more deeply. Therefore, we argue that even though achieving an effective positive detection performance is important, tracking and visualising the development of the mental disorder is equally relevant~\citep{Rissola:2020}. In fact, an accurate detection system can be more useful if it provides a way of understanding the factors that lead to a certain decision. As argued by ~\cite{Walsh:2020}, this issue, along with other factors, might prevent most of the risk-assessment and decision-making technologies from ever been used in real-life settings.


\par Bearing this in mind, we believe it is of great importance to conduct experiments providing insights on how individuals' language attributes and online behaviour are distinctive among users affected by mental health concerns as well as between different mental disorders. We focus on several language dimensions as well as on the social engagement and posting trends of social media users. We especially emphasise the importance of finding innovative ways to visualise the onset and development of a mental disorder. In this way, forecasting systems could be complemented with preliminary step-by-step directions for practitioners to identify high-risk individuals based on both statistical and visual analyses. Furthermore, it is essential for practitioners to count on a way to assess the degree of the differences identified between individuals; hence visual outcomes should not be presented as standalone objects but supported with the significance that the observed trends have.

\par In this work, we conduct a thorough study of various dimensions of language and online behavioural traits to characterise users affected by mental disorders on two popular social media platforms, namely, Reddit and Twitter. Moreover, we present several methods for visualising the data to give mental health professionals a deeper understanding of the information extracted. To this aim, we start by comparing users affected by a particular disorder against control individuals in each platform separately. Subsequently, our second aim is to determine whether different mental disorders share the same features or if they are distinct according to the dimensions studied. Finally, we investigate whether the trends and patterns are held on both social media platforms in order to identify and quantify potential differences and better understand which analysis techniques are more suitable for each case.
\par Our main research questions related to the language and behaviour of users of social media platforms are:
\begin{itemize}
    \item \textbf{RQ1}: How different is the language of users with mental health disorders compared to control individuals?
    
    \item \textbf{RQ2}: How do social engagement and posting trends of users with mental health disorders differ from control individuals?
    
    \item \textbf{RQ3}: To what extent are the language and online behavioural traits of various mental disorders different?
    
    \item \textbf{RQ4}: How can language-specific and emotional information be visualised for mental health practitioners during the diagnosis process?
    
    \item \textbf{RQ5}: How can users' engagement in relationship with their mental health condition be visualised as it develops?
    
    \item \textbf{RQ6}: How different are the language and online behavioural traits of users with mental health disorders among different social media platforms? In this respect, how much are the constraints imposed by social media influencing such users?

\end{itemize}


\par This paper extends our previous work on improving the understanding of mental disorders on online social media~\citep{Rissola:2020}. We previously stressed that language technologies have a great potential to identify and exploit latent linguistic nuances encoded in users’ textual records that correlate with mental disorders; thus, they can provide practitioners with a holistic and comprehensive way to assess the individual. Driven by this, we conducted an extensive language analysis to improve the understanding and visualisation of the features that characterise and differentiate social media users affected by mental disorders on Reddit. In particular, we proposed a series of techniques based on statistical and visual analyses which are able to identify significant differences in the language of high-risk individuals on Reddit.

\par With this in mind, we further explore how mental disorders are manifested on social media as we extend the language analyses from Reddit to Twitter, another popular platform subjected to strong constraints, such as the length of a posting in number of characters. Furthermore, we complement the language dimensions studied on both platforms with a behaviour analysis. In particular, we study a set of attributes that profile the online behavioural traits and trends of affected users, such as the posting regularity.  
As shown later, the differences observed, mostly between positive and control instances, have a varied magnitude depending on the social media platform. Therefore, given the characteristics and limitations imposed by the platform, there are attributes that provide stronger clues about the manifestation of mental disorders. For instance, many of the language dimensions analysed in this work exhibit larger variance on Reddit than on Twitter.
This is an ideal complement to our previous work on improving the understanding of how mental health disorders are manifested on social media as it uncovers the differences in people's language and behaviour based on their participation on social media platforms with different characteristics and constraints. In summary, this paper extends our previous work as follows:


\begin{itemize}
    \item It extends the language closed-vocabulary analysis adding a complementary dictionary-based tool known as Empath~\citep{Fast:2016} and compares the new results with the previous ones.
    \item It complements language analyses on Reddit by profiling users' online behavioural traits, such as the usage of elements like hashtags, mentions and emoticons.
    \item It extends Reddit's language and online behaviour analyses to another social media platform, in particular, to Twitter data.
    \item It presents a thorough comparison between the language and online behaviour of users on both social media platforms to shed light on how distinctive the expression of mental disorders could be on different platforms to understand which of the dimensions analysed are more suitable for each case.
\end{itemize}

\par To the best of our knowledge, this is the first time that a study of mental health disorders on social media is analysed in depth based on two platforms with highly diverse characteristics. This analysis is a valuable step towards developing better risk-assessment and decision-support tools.


\par Outcomes from our thorough analysis showed that psychometric attributes, emotional expression and social engagement provide a quantifiable and significant way to differentiate individuals affected by mental disorders from healthy ones. Conversely, comparing the mental disorders analysed, on Reddit and Twitter respectively, we were not able to find significant attributes which could differentiate one from the other, though such disorders are identified in the language and behaviour of the individuals. In addition, we observe less quantifiable differences between positive and control groups on Twitter than on Reddit, suggesting that the use of language on Twitter is less distinguishable between users potentially suffering mental disorders. This could be due to the platform constraints, most importantly the character limit. Nonetheless, the usage of social engagements elements, such as hashtags, mentions and emojis, revealed meaningful differences between the various groups on Twitter.

\par The remainder of the paper is organised as follows. Section \ref{sec:related_word} summarises the related work; Section \ref{sec:approach} details the approach followed to answer the research questions; Section \ref{sec:dataset} describes the specific data used in this work; Section \ref{sec:results_analyses} presents the corresponding results and analyses; Section \ref{sec:conclusions_future_work} concludes the work.

\section{Related Work} 
\label{sec:related_word}
Social media has been extensively studied for its potential to automatically identify individuals suffering from mental disorders. Here, we summarise a number of relevant studies that have attempted to examine the connection between language, behaviour and mental disorders on social media from an analytical perspective.

\par Using online social media data as a gauge of depression, ~\cite{Park:2012} performed a preliminary study to identify whether this data was truly representative of users' clinical depressive symptoms. The data analysis conducted revealed that depression was mentioned most frequently for describing one's depressed state and, to less degree, for sharing general information about depression.




\par An early work on automatic depression detection was undertook by ~\cite{Choudhury:2013a} who
gathered ground truth data on the presence of depression using crowd-sourcing. Their data analysis revealed that depression sufferers exhibited a noticeable decrease in social activity (lower posting volume), greater negative emotion and reduced arousal, in addition to higher usage of first-person pronouns when compared with non-depressed users.  


\par ~\cite{Coppersmith:2015a} investigated Twitter data to determine which language features could be helpful to distinguish between mental disorder sufferers and healthy individuals. It was found that, though a set of useful features could be identified, there are still language differences in communicating about mental health concerns which worth investigate.


\par ~\cite{Reece:2017} conducted a state-space temporal analysis whose goal was to track the evolution of depression and PTSD. They developed time series models which, exploiting posts' textual content, presented distinctive patterns between positive and controls users without any supervision signal. Furthermore, such models exhibit their potential to outline a plausible timeline for the disorders development (onset and recovery).


\par ~\cite{DeChoudhury:2016} studied several mental health and suicide support communities on Reddit\footnote{Denominated \textit{subreddits} on the platform.} whose members mostly participate looking for help and support. The authors identified that transitions from mental health disclosure to an expression of suicidal ideation were associated with various markers, such as heightened self-attentional focus, poor linguistic coherence, reduced social engagement, and expression of negative affects like as anxiety. 

\par ~\cite{Gkotsis:2016} analysed various mental health subreddits to discover discriminating language features between the users in the different communities. Their study concluded that, in general, non-topically related subreddits had distinctive syntactic and lexical features and condition-specific vocabularies.
Similarly, ~\cite{Gaur:2018} presented an unsupervised approach to map the content of various mental health-related subreddits to the best matching DSM-5\footnote{DSM-5 stands for \textit{Diagnostic and Statistical Manual of Mental Disorders - 5th Edition.}} categories. They developed a domain-specific lexicon alongside a drug abuse ontology and used them to quantify the relationship between subreddits' content and DSM-5 categories by automatically assigning the corresponding labels. 
Also on Reddit, ~\cite{Yoo:2019} studied how users perceived bipolar and depressive disorders and shared medical information about them on the respective mental health support communities on the platform. Relying on semantic network analysis, they observed that, overall, users on both communities often expressed negative emotions, and showed great interest in writing about their past episodes and sleeping problems.

\par Although insightful, such studies of Reddit communities might not provide accurate information about the distinction between mental health sufferers and healthy users as several members of such subreddits are persons who search for advise or to share their experience about close friends or relatives who could be experiencing a mental health concern. In this work, we focus on the analysis of users' language attributes irrespective of the topic discussed.




Overall, much work is focused on automatically identifying whether online social media users are deemed as positive mental health concerns cases. On very few occasions, the authors of such works attempt to obtain some explanations from the outcomes of the classification models to better understand why certain decisions were taken, as happened with some eRisk participants~\citep{Amini:2020,Uban:2021,Aragon:2021}. However, research devoted to understanding, measuring, visualising and providing insight on the attributes that characterise such users and differentiate them both from healthy individuals and between diverse disorders has been noticeably scarce. Moreover, the language dimensions and behavioural traits studied in this work have not yet been contrasted on social media platforms of different characteristics and constraints.
\section{Experimental Design}
\label{sec:approach}
Language and behaviour analysis usually focus on one of two applications, \textit{prediction} (typically, involving supervised machine learning algorithms) or \textit{insight}. In prediction, the ultimate goal is to accurately map language and behaviour into a single or few given outcomes. Insight is conceptually focused on the opposite, finding the language and behaviour that is most characteristic of an outcome, i.e., identifying the patterns that provide a window into thoughts, attitudes, psychology or health of people. In this work, we sought to gain \textit{insights} on the manifestation of mental disorders on online settings, particularly social media platforms. With this in mind, in the following section we outline the experimental design we defined to analyse social media posts from the perspective of language and behaviour. We describe the main outcome of each experiment and state how the differences between the classes under analysis can be quantified.


\subsection{Open-Vocabulary}
\label{subsec:approach_open_vocab}
\partitle{Vocabulary Uniqueness: }To address \textbf{RQ1} and \textbf{RQ6}, we analyse the similarity (or diversity) of the unique sets of words that comprise the vocabulary of the classes (i.e., positive and control). With this analysis we can investigate to which extent the two groups share a common vocabulary and whether specific words (if any) could be only employed by the members of a certain class.
\par Considering each vocabulary as a set, we inspect the relative size of their intersection by computing Jaccard's index. This metric provides a way to quantify the similarity between finite sample sets. Formally, let $P$ be the unique set of words obtained from positive users, e.g. self-harm, and $C$ be the unique set of words obtained from control users. We calculate Jaccard's index as follows:
\begin{equation}
J(P, C) = |P \cap C |/|P \cup C|~.    
\end{equation}
\noindent As we see, the index gives us the ratio of the size of the intersection of $P$ and $C$ to the size of their union. As the value approaches $1$ the sets share an increasing overlap of elements. Conversely, an index close to $0$ indicates that the sets are more diverse among themselves.

\partitle{Word Usage: }Complementary to analysing vocabularies overlapping and differences, we also study word usage patterns as a mean to inspect the language of the different classes. Here, we sought to address \textbf{RQ1}, \textbf{RQ3}, \textbf{RQ4}, \textbf{RQ6} by computing and comparing the language models for each class. This analysis aims to quantify the differences that might emerge between the classes in terms of the probability of using certain words more than others.



\par Based on both current and historical data, language models are probability distributions over the words in a collection of documents which intend to capture the regularities of language. Thus, when considering a unigram language model each of the terms in the vocabulary will receive a probability based on its occurrence patterns. An estimate of the likelihood that a word will appear in a document $D$ in a collection of documents $S$ is done using:

\begin{equation}
\label{eq:language_model}
P(w_i|D) = (1 - \alpha_D)P(w_i|D) + \alpha_DP(w_i|S)~,    
\end{equation}


\noindent where $\alpha$ is a smoothing coefficient used to control the probability assigned to out-of-vocabulary words. In particular, we use the linear interpolation method, known as Jelinek-Mercer smoothing, where $\alpha_D=\lambda$, i.e., a constant. To estimate the probability for word $w_i$ in the collection we use $s_{w_i}/|S|$, where $s_{w_i}$ is the number of times a word occurs in the collection, and $|S|$ is the total number of words occurrences in the collection. In this work, $D$ identifies all the documents in a specific class, i.e., we concatenate all the documents of a particular class, such as self-harm. While $S$ is the union of all the documents of two classes in a corpus, i.e., positive and control.

\par After calculating the language models for each class, we graphically analyse the probability distributions computed and study how much the distributions diverge. To support our visual observations, we use Kullback-Leibler divergence (KL). This measure from probability and information theories allows us to quantitatively measure the extent to which the probabilities distributions under observation differ. When the distributions compared are identical, their KL-divergence reaches $0$. Thus, the divergence is always positive and is larger for distributions that are more different. Given the \textit{true} probability distribution $P$ and control distribution $C$, the KL-divergence is defined as:


\begin{equation}
KL(P||C) = \sum_{x}^{} P(x) log \frac{P(x)}{C(x)}~.    
\end{equation}

\subsection{Linguistic Style and Psychometric Attributes}
A usual approach in which language can be associated to psychological variables consists in counting words that are included within different categories of language manually defined~\citep{Choudhury:2013a,Coppersmith:2015a,Chung:2007}. As opposed to the experiment outlined in the previous section, this type of analysis is called ``close-vocabulary''~\citep{Schwartz:2013}.
By studying \textit{function words}\footnote{Function words contribute to the structure of the sentence, rather than to its meaning.} and topic-specific vocabulary, we address \textbf{RQ1}, \textbf{RQ3}, \textbf{RQ4}, and \textbf{RQ6}. This kind of analysis is aimed at quantifying specific stylistic patterns that could distinguish positive cases of mental disorders from control cases. For instance, when people are depressed, they tend to focus on themselves, and therefore, it is expected that they would employ personal pronouns like ``I'' more frequently~\citep{APA:2016}. Additionally, individuals suffering from certain mental health disorders may use more words related to specific topics. We showcase this later by comparing depression and self-harm to cases of anorexia.

\par It worth noting that we decide to keep the stop-words since many words such as pronouns, articles and prepositions disclose individuals' way of thinking, emotional state, personality, and relationship with others~\citep{Chung:2007}. Indeed, such words, called function words, represent less than 0.1\% of people's vocabulary, but nearly 60\% of the words they employ~\citep{Chung:2007}.

\par By analysing the writing style of individuals, \textit{Linguistic Inquiry and Word Count}~\citep{Tausczik:2009} (LIWC)\footnote{See \url{http://liwc.wpengine.com/}} provides mental health practitioners with quantitative data regarding those individuals mental health. LIWC consists of an extensive set of dictionaries compiled by psychologists, and including psychologically meaningful categories, which are useful for analysing the linguistic style patterns of an individual's writing style. In this research, we determine the proportion of documents authored by each individual that has at least one word on specific LIWC categories and produce a distribution. Specifically, we select a subset of the psychometric categories where significant differences were found between positive and control group members. Afterwards, we depict and compare the resulting distributions employing box-plots.


\par Furthermore, we extend this analysis using Empath\footnote{See \url{http://empath.stanford.edu/}}~\citep{Fast:2016}. This text analysis tool shares a high correlation with gold standard lexicons, such as LIWC, yet it covers a broader and dynamic set of emotional and topical categories. Conversely to hand-tuned dictionary-based tools, Empath's data-driven and human-validated categories are derived from existing knowledge bases and literature on human emotions available on the web by means of unsupervised language modelling. Similarly to the analysis conducted with LIWC, we use box-plots to visualise the distributions obtained from a subset of categories where we observed significant differences between control and positive users and contrast them.

\subsection{Emotional Expression}
Typically, people express their feelings, attitudes and emotions through their words. As an example, \textit{gloomy} and \textit{cry} evoke sadness, while \textit{delightful} and \textit{yummy} denote joy. In this section, we focus on \textbf{RQ1}, \textbf{RQ3}, \textbf{RQ4}, and \textbf{RQ6} by examining how people with mental disorders convey their emotions in social media posts. In addition, we study whether emotional expression could be a differential factor between users affected by mental health concerns and those who are not. To this aim, we followed Plutchik's model of emotions ~\citep{Plutchik:1980} which proposes the existence of eight basic and prototypical emotions, these include joy, sadness, anger, fear, disgust, surprise, trust and anticipation. In addition, we also considered the polarity (positive or negative).
\par To capture word-emotion connotations, we apply the emotion lexicons that ~\cite{Mohammad:2013,Mohammad:2018} developed, in which each word is associated with the emotion it conveys. These lexicons include commonly used words from social media platforms along with common English terms. Furthermore, they contain words that not necessarily evoke a certain emotion, but which co-occur with those that do. As an example, the words \textit{death} and \textit{failure} refer to notions which are often associated to sadness, and, hence, suggest some kind of sadness.


\subsection{Social Engagement and Posting Trends}
~\cite{Coppersmith:2014b} suggested that users' online behavioural traits could be profiled by studying various attributes related to users' documents, such as creation time and length, as well as users' level of participation on the social media platforms. Such features constitute proxies, albeit imperfect, for significant findings in the mental health literature that may be manifested and quantified on online settings. Furthermore, on some occasions they might provide a starting point for further research into behavioural traits as observed on social media. As an example, ~\cite{Reece:2017} found a relationship between higher average document length and both depression and PTSD. Yet, in the traditional medical literature few studies link verbosity to mental health concerns. With this in mind, we address \textbf{RQ2} and \textbf{RQ6} by profiling the online behaviour of users considering the following elements:


\begin{itemize}
    \item Mentions: A document containing another account's username, preceded by the ``@'' symbol. For instance, \texttt{Hello @earissola!}.
    \item Hashtags: Unbroken word or phrase preceded by ``\#'' on Twitter, when a hashtag is used in a tweet, it becomes linked to all of the other tweets that include it. For instance, \texttt{\#WishYouWereHere}.
    \item All-Caps: Words completely written in capitals. Usually, when uppercase letters are used in emails, text messages, and social media, they are intended to emphasise as if the individual was speaking in an assertive voice or shouting to convey an emotion such as anger or dismay.
    \item ASCII emoticons: Emoticons written as plain text, such as \texttt{>:(}, \texttt{:-P}, \texttt{<3}.
    \item Emojis: A smiley or an ideogram inserted into an e-mail or web page. Emojis are similar to emoticons, though they are pictures rather than typographical representations.
    \item Other annotated elements: It includes words with emphasis (such as \texttt{I don't *think* I...} and \texttt{a *great* time}), censored words (such as \texttt{f**k}, \texttt{s**t}) and repeated words.
    \item Retweets (RT): On Twitter, usually ``RT'' indicates that someone is re-posting someone else's content, i.e., quoting another tweet.
    \item Submission type: On Reddit, users submit content in the form of posts and comments. While posts are used to start an online conversation (called a \textit{thread}), comments are nested responses to other comments or posts.
\end{itemize}

\par We examine each user's documents to determine what percentage scores positively on any of these elements (i.e., have at least one occurrence of that element). Using box-plots, we visualise and contrast the distributions obtained.

\par Finally, we study the time-gap between two consecutive documents for each user in the collection to address \textbf{RQ5} and \textbf{RQ6}. By analysing the mean and variance of this variable on a monthly basis and aggregating it according to the respective group (e.g., PTSD), we analyse and compare the posting regularity of each class. In this way, we can study how  users’ engagement on the social media platform develops in relationship with their mental health condition. Formally, let $t_1, t_2, \dots, t_n$ be the timestamps of the elements that comprise the chronology of documents of a user. The average document time-gap for a user is computed as:
\begin{equation}
\frac{1}{n-1} \sum_{t=2}^{n-1} t_{n-1} - t_n,     
\end{equation}
where $n$ represents the number of documents that a user has written. Once we have computed the time-gap for each user, we aggregate the values obtained according to the class and month of the year and plot the distribution.

\section{Data}
\label{sec:dataset}
In this work, we hypothesise that users with a mental disorder behave differently on different social media platforms. One approach, data type, and data modality does not reveal the same information on every social media platform. For example, language use can be a very discriminating feature on a platform, while performing poorly on another. For this reason, we study various mental disorder indicators and visualisation techniques on two mainstream social media platforms, namely, Reddit and Twitter. Our choice is motivated by the fact that the two platforms have very different environment and features, leading to a very different range of behaviours by their users~\citep{Roberts:2012}.

\par \textbf{Reddit} ``is a social news aggregation, web content rating, and discussion website, recently including livestream content through Reddit Public Access Network''\footnote{See \url{https://www.reddit.com}}. It consists of various \textit{subreddits}, each of which focuses on a different topic. Users can post pictures, web links, or other types of content. The platform does not enforce any extreme restrictions in terms of the length of posts.
\textbf{Twitter} ``is an American microblogging and social networking service on which users post and interact with messages known as tweets''\footnote{See \url{https://twitter.com}}. As the description suggests, users on Twitter post short messages. The structure and features of Twitter are very different from Reddit. For instance, sharing a post (i.e., retweet) is a common means of interaction with a post. Limiting each post to 280 characters has various effects on the way people express themselves and their feelings. It can also affect the topic and frequency of the posts on Twitter instead of other platforms~\citep{Gligoric:2018}. 

With this in mind, as a test bed for our experiments we use the datasets developed at CLPsych~\citep{Coppersmith:2015a} (Twitter) and eRisk~\citep{Losada:2018,Losada:2019} (Reddit) experimental frameworks. These initiatives sough to develop a common evaluation framework for researchers interested in advancing the state of the art on the assessment of mental disorders, such as depression or PTSD, on social media platforms.
Each dataset is comprised of submissions in the form of posts created by users of social media platforms, namely Reddit and Twitter. Two groups of users are included in each dataset:
\begin{inparaenum}[(a)]
    \item cases of users potentially suffering from mental health concerns, such as depression (positive group); and
	\item control individuals.
\end{inparaenum}
On both experimental frameworks individuals in the positive group were collected by searching for self-expressions of diagnoses (e.g., the sentence ``I was diagnosed with depression'') and manually verifying that the posts retrieved truly contained a genuine statement of diagnosis~\citep{Coppersmith:2014a}. Conversely, members of the control group were randomly sampled from the extensive pool of users available on the platform. Given the API limits imposed by each social media platform, the datasets creators were able to collect up to $2,000$ submissions per user for eRisk and $3,000$ for CLPysch.


\par We decide to conduct our various analysis using these collections since they have been carefully curated, validated and used through the various experimental frameworks' editions. The fact that they have been extensively tested by several researchers allow us to build our work based on their experience with the collections. Furthermore, they are publicly available for research, which is not usually the case in the domain under study. It should be noted that a direct comparison at the level of classes between eRisk and CLPsych is limited by the availability of mental disorders that were included in each dataset. The only overlapping label between both platforms is depression. Nonetheless, we consider both datasets as valuable and valid assets to perform the analyses proposed in this work as we are interested in determining whether the general trends and patterns we observed on users affected by mental disorders are held on both platforms, despite the particularities that each one has. Especially when users are subject to various types of constraints (e.g.~character limit), which could influence their writing style and expression~\citep{Gligori:2019}.
Table \ref{tab:dataset} presents a summary of the datasets employed in this research.

\begin{table*}[ht]
	\scriptsize
	\setlength{\tabcolsep}{3pt}
	\caption{Summary of eRisk 2018 and 2019 collections (Reddit) and CLPysch 2015 (Twitter). The activity period represents the number of days passed from the first to last the document collected for each user.}
	\begin{center}
	    \begin{tabular}{c}
	        (a) Reddit (eRisk) \\
	         \begin{tabular}{lrrrrrrrr} 
			\toprule 
			& \multicolumn{2}{c}{Depression} & \phantom{abc} & \multicolumn{2}{c}{Anorexia} & \phantom{abc}  & \multicolumn{2}{c}{Self-Harm} \\
			\cmidrule{2-3} \cmidrule{5-6} \cmidrule{8-9}
			& Positive &  Control  & & Positive & Control & & Positive &  Control  \\
			\midrule\midrule
			\# of Users & $214$ & $1,493$  & & $61$ & $411$ && $41$ & $299$ \\
			\# of Documents & $89,999$ & $982,747$  & & $24,776$ & $227,219$ & & $7,141$ & $161,886$ \\
			Avg. \# of Documents/User & $420.5$ & $658.2$  & & $406.16$ & $552.84$ & & $174.17$ & $541.42$ \\
			Avg. \# Words/Document & $45.0$ & $35.3$  & & $64.6$ & $31.4$ &  & $39.3$ & $28.9$ \\
            Avg. Activity Period (Days) & $\approx 658 $ & $\approx 661$  & & $\approx 799$ & $\approx 654$ &  & $\approx 504$ & $\approx 785$ \\
			\bottomrule
		\end{tabular} 
		\\
		\\
		(b) Twitter (CLPsych) \\
		
		\begin{tabular}{lrrrrr} 
			\toprule 
			& Depression  & & PTSD & & Control  \\
			\midrule\midrule
			\# of Users & $477$ && $396$ && $872$ \\
			\# of Documents & $1,131,997$ & & $919,131$ & & $1,978,121$  \\
			Avg. \# of Documents/User & $2373.73$  && $2321.64$ && $2268.51$\\
			Avg. \# Words/Document & $13.9$  && $16.5$ && $13.8$ \\
            Avg. Activity Period (Days) & $\approx 379$ & & $\approx 479$  &  & $\approx 460$ \\
			\bottomrule
		\end{tabular}
	          
	    \end{tabular}
		
	\end{center}
	\label{tab:dataset}
\end{table*}

\section{Results and Analyses}
\label{sec:results_analyses}
In this section, we outline the results reached after analysing the various collections presented in Section~\ref{sec:dataset} which were used for the experiments described in Section~\ref{sec:approach} aiming at addressing the research questions posed in the introduction.




\subsection{Open-Vocabulary}
\partitle{Vocabulary Uniqueness:} In Table \ref{tab:vocab_cmp} we analyse the vocabulary used by different groups of individuals suffering from mental health concerns (i.e, depression, anorexia and self-harm for Reddit and depression and PTSD for Twitter) in comparison to the respective control groups. In particular, we study vocabularies' difference, intersection and union. Our analysis of Reddit data indicates that positive cases of depression and anorexia are highly similar to the relevant control groups, with Jaccard indexes of $59\%$ and $65\%$, respectively.
The self-harm cases, however, tend to use a broader range of words in their documents as compared to the control group ($44\%$). One possible explanation for such difference might be related to the way in which the collections where built. When creating depression and anorexia collections, eRisk's organisers also included in the respective control groups users who were active on Reddit's communities devoted to discussing about depression or anorexia, but were not suffering from any disorder. Examples of this are a mental health practitioner giving support to other forum members or people who have some relative suffering from depression or anorexia. This type of control users made the task (i.e., determining early signs of depression or anorexia) more challenging because their topics of interest are highly related to those of the positive group. Conversely, none of these users were included in self-harm's control group.

\par On Twitter, differences between positive and control groups are considerably larger. A Jaccard index of $26\%$ for depression and $27\%$ for PTSD indicates a more distinctive vocabulary. In spite of this, it should be noted that such noticeable differences may be due in part to the informal nature that characterises microblogging platforms~\citep{Choi:2012,Metzler:2012,Nepomnyachiy:2014} along with Twitter's limit of maximum characters per post. Therefore, abbreviations, compound words hashtags, mentions, internet slangs and misspellings are common, in many cases deliberate, causing an almost linear growth of the vocabulary size~\citep{Rissola:2016}. In fact, the number of unique terms on Twitter is substantially larger than on Reddit, possibly causing a smaller overlap between the sets compared.
\par Furthermore, such comparisons between the different vocabularies allow us to outline which words are used by positive groups and not by control users. For instance, on Reddit we found the following terms to be distinctive for positive groups:
self-harm, trazodone\footnote{\textit{Trazodone} is a medication that helps treat depression.} (Depression); anorexics, depersonalization\footnote{People suffering from depersonalisation feel detached or disconnected from their bodies and thoughts.}, emetrol, peptol\footnote{\textit{Peptol} and \textit{Emetrol} are pharmacological drugs that relieve stomach discomfort.} (Anorexia).

\begin{table*}[ht] 
	\scriptsize
	\setlength{\tabcolsep}{3pt}
	\caption{Vocabularies comparison between positive and respective control users. KL-divergence computed across the language models obtained for the documents of positive and control users. As a reference, the KL-Divergence is also calculated between the different control groups on Reddit. }
		\begin{center}
		(a) Reddit (eRisk) \\
		\begin{tabular}{lrrrrrrrr}
			\toprule 
			& \multicolumn{2}{c}{Depression} & \phantom{abc} & \multicolumn{2}{c}{Anorexia} & \phantom{abc}  & \multicolumn{2}{c}{Self-Harm} \\
			\midrule\midrule
            \# of Unique Words Positive & \multicolumn{2}{c}{$41,986$} & \phantom{abc} & \multicolumn{2}{c}{$21,448$} & \phantom{abc}  & \multicolumn{2}{c}{$11,324$} \\
            \# of Unique Words Control & \multicolumn{2}{c}{$70,229$} & \phantom{abc} & \multicolumn{2}{c}{$31,980$} & \phantom{abc}  & \multicolumn{2}{c}{$25,091$} \\\hdashline
			Jaccard's Index (Positive vs. Control) & \multicolumn{2}{c}{$0.59$} & \phantom{abc} & \multicolumn{2}{c}{$0.65$} & \phantom{abc}  & \multicolumn{2}{c}{$0.44$} \\
			Difference Size (Positive vs. Control) & \multicolumn{2}{c}{$218$} & \phantom{abc} & \multicolumn{2}{c}{$229$} & \phantom{abc}  & \multicolumn{2}{c}{$49$} \\
			Difference Size (Control vs. Positive) & \multicolumn{2}{c}{$28,461$} & \phantom{abc} & \multicolumn{2}{c}{$10,761$} & \phantom{abc}  & \multicolumn{2}{c}{$13,816$} \\\hdashline
			KL(Positive$||$Control) & \multicolumn{2}{c}{$0.18$} & \phantom{abc} & \multicolumn{2}{c}{$0.18$} & \phantom{abc}  & \multicolumn{2}{c}{$0.18$} \\
			KL(Control$||$Positive) & \multicolumn{2}{c}{$0.21$} & \phantom{abc} & \multicolumn{2}{c}{$0.31$} & \phantom{abc}  & \multicolumn{2}{c}{$0.20$} \\
			KL(Control$||$Control) & \multicolumn{2}{c}{$0.08$} & \phantom{abc} & \multicolumn{2}{c}{$0.07$} & \phantom{abc}  & \multicolumn{2}{c}{$0.10$} \\
			\bottomrule
		\end{tabular}
	\end{center}
	\label{tab:vocab_cmp}

		\begin{center}
		(b) Twitter (CLPsych) \\
		\begin{tabular}{lrrrrr} 
			\toprule 
			& \multicolumn{2}{c}{Depression} & \phantom{abc} & \multicolumn{2}{c}{PTSD} \\
			\midrule\midrule
            \# of Unique Words Positive & \multicolumn{2}{c}{$150,508$} & \phantom{abc} & \multicolumn{2}{c}{$149,393$} \\
            \# of Unique Words Control & \multicolumn{2}{c}{$238,712$} & \phantom{abc} & \multicolumn{2}{c}{$238,712$}  \\\hdashline
			Jaccard's Index (Positive vs. Control) & \multicolumn{2}{c}{$0.26$} & \phantom{abc} & \multicolumn{2}{c}{$0.27$}  \\
			Difference Size (Positive vs. Control) & \multicolumn{2}{c}{$70,492$} & \phantom{abc} & \multicolumn{2}{c}{$66,945$}  \\
			Difference Size (Control vs. Positive) & \multicolumn{2}{c}{$158,696$} & \phantom{abc} & \multicolumn{2}{c}{$156,264$} \\\hdashline
			KL(Positive$||$Control) & \multicolumn{2}{c}{$0.08$} & \phantom{abc} & \multicolumn{2}{c}{$0.09$} \\
			KL(Control$||$Positive) & \multicolumn{2}{c}{$0.09$} & \phantom{abc} & \multicolumn{2}{c}{$0.09$} \\
			\bottomrule
		\end{tabular}
	\end{center}
	\label{tab:vocab_cmp}
\end{table*}

\begin{figure}[!ht]
    \centering
    \subfigure[][Reddit: Depression vs. control]
    {
        \includegraphics[width=0.47\textwidth]{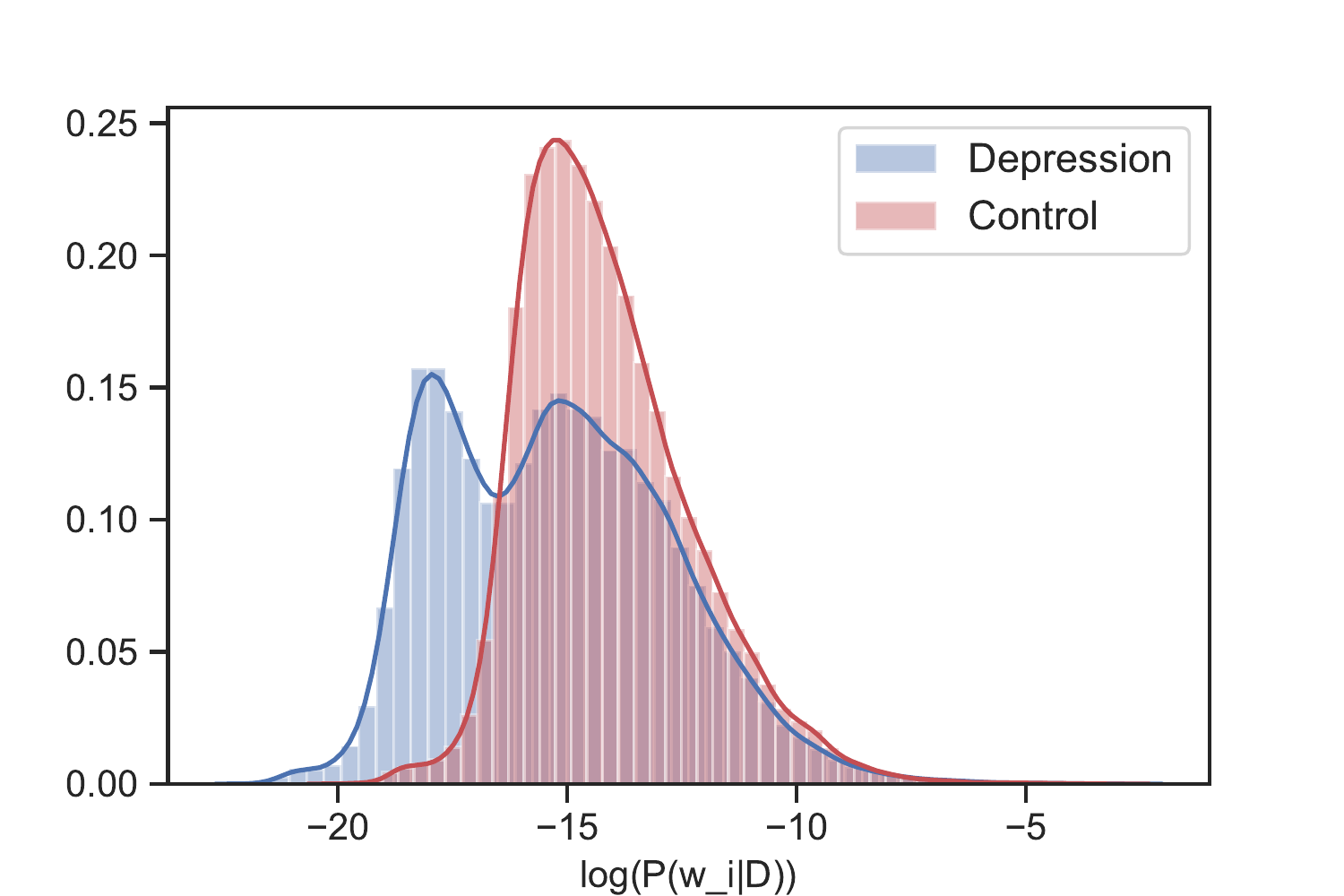}
        \label{fig:lm_jm_dep_ctrl_erisk}
    }
    \subfigure[][Reddit: Anorexia vs. depression vs. \newline \hspace*{1.5em} self-harm]
    {
        \includegraphics[width=0.47\textwidth]{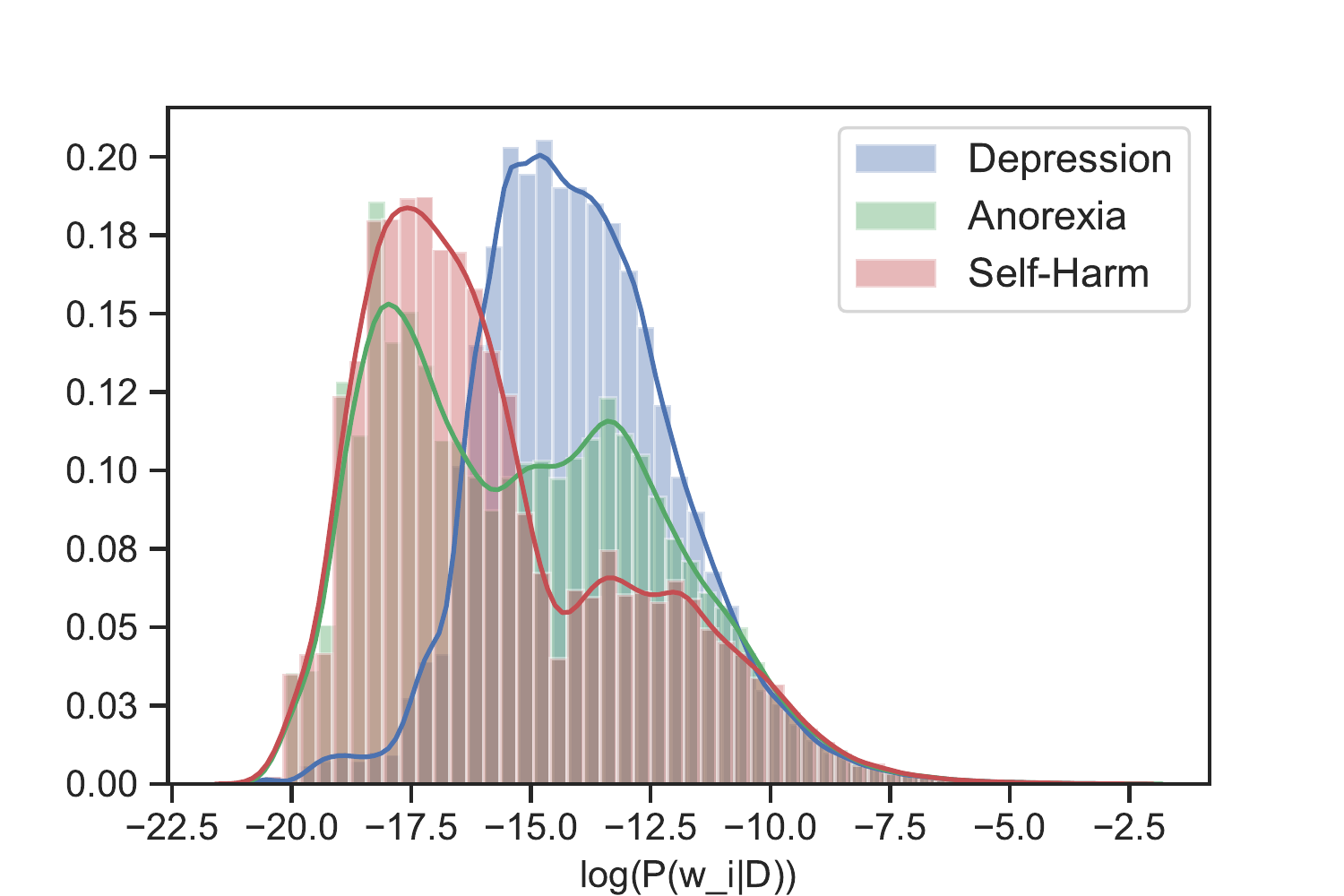}
        \label{fig:lm_jm_all_posv}
    }
        \subfigure[][Twitter: Depression vs. control]
    {
        \includegraphics[width=0.47\textwidth]{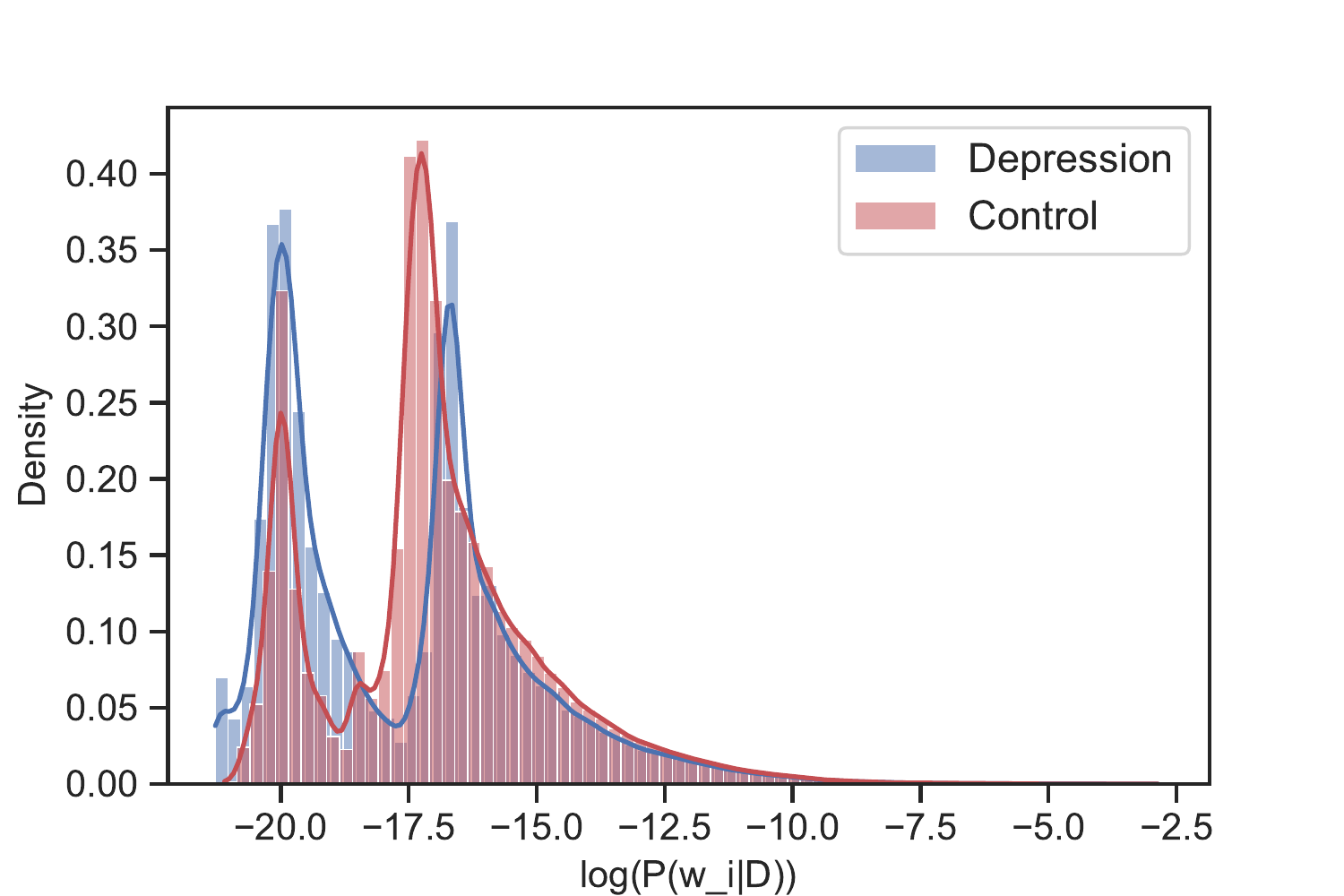}
        \label{fig:lm_jm_dep_ctrl_clpsych}
    }
    \subfigure[][Twitter: PTSD vs. control]
    {
        \includegraphics[width=0.47\textwidth]{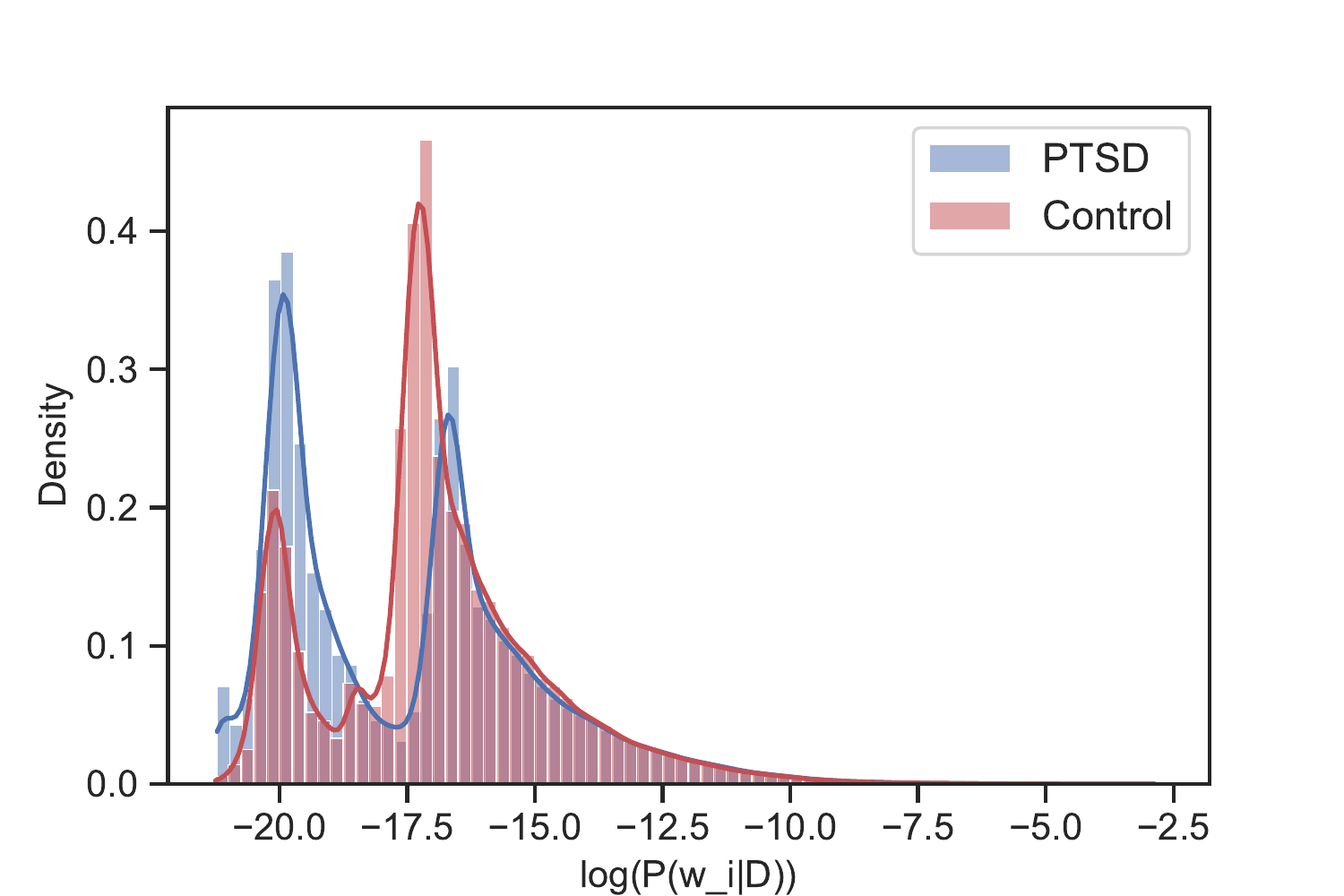}
        \label{fig:lm_jm_ptsd_ctrl_clpsych}
    }
    \caption {Language models probability distribution comparison on Reddit and Twitter (best viewed in colour). }
    \label{fig:lm}
\end{figure}


\partitle{Word Usage: }In Figure \ref{fig:lm} we examine and contrast the language models derived from the various collections and classes. As previously observed, there are cases of out-of-vocabulary words between the classes. That is, there are terms which only appear in the positive class vocabulary and not in the control one (and vice-versa). In that case, smoothing becomes necessary to control the probability assigned to such out-of-vocabulary words.
To illustrate how different the language use between positive and controls users could be, we turn to Figure \ref{fig:lm_jm_dep_ctrl_erisk} in which we showcase the language models of depression and control classes on Reddit. As observed, word usage between positive and control depicts very different trends. In contrast, as Figures \ref{fig:lm_jm_dep_ctrl_clpsych} and \ref{fig:lm_jm_ptsd_ctrl_clpsych} reveal, on Twitter such differences are much smoother as there is a notable overlap between the distributions.


\par Table \ref{tab:vocab_cmp} presents the KL-divergence computed between the positive and control groups' language models on both platforms. Since the KL-divergence lacks an upper bound, we also compute it among the different control groups to serve as point of reference. We observe that KL-divergence calculation reinforces the trends observed in the plots, specially on Reddit. As we contrast the distribution of different control groups on Reddit, we note smaller KL-divergence values ($\approx 0$), denoting that these distributions are very similar. Conversely, this value is considerable larger when comparing positive and control classes.
Moreover, when compared to Reddit, the positive groups on Twitter exhibit almost half of the KL-divergence with respect to controls. Interestingly, the divergence observed on Twitter between positive and controls groups and vice-versa is practically equal. Recall that KL-divergence is a distribution-wise asymmetric measure that allows to exactly calculating how much information is lost when we approximate one distribution with another. Therefore, the fact that this difference appears to be symmetric, emphasises the similarity observed between the distributions.



\par Finally, as observed in Figure \ref{fig:lm_jm_all_posv}, it is difficult to spotlight prominent differences between the language models of the three positive classes on Reddit. Anorexia and depression exhibit similar word usage distributions. In comparison with either depression or anorexia, self-harm shows the most noticeable difference. On Twitter, positive group language models present an even larger overlap\footnote{Given that the figure is not providing any observable trend and to avoid overloading the article, we decided not to include it.}.
This strengthen the fact that the word probability distribution between the various positive groups is very similar and how they use words.
Notice that language models for depression in Figures \ref{fig:lm_jm_all_posv} and \ref{fig:lm_jm_all_posv} exhibit different shapes. The reason for this is related to they way language models are computed. As stated in Section \ref{sec:approach} an estimate of the likelihood for each word is calculated based on its frequency on both the concatenation of a class of documents and the whole collection. Thus, depending on the classes involved in the computation, the second term in formula \ref{eq:language_model} will vary. Recall that this term allows to deal with out-of-vocabulary words.

\par Practitioners could used the open-vocabulary analysis presented in this section as supporting material to paint a bigger picture about users mental health state. For instance, the language model of a potentially affected user could be contrasted against reference language models for positive and control classes to determine the which one it resembles the most. we have randomly sampled 10\% of users with anorexia and 10\% of control users and calculated the KL-divergence of each individual language model against reference language models for positive and control, and report the average values of each class. We observe that users with anorexia consistently exhibit a lower divergence with respect to positive reference language model than to the control one (0.41 vs.~0.49). Conversely, control users showed a higher similarity with the control reference language model than with the positive one (0.82 vs.~0.90).

More precisely, we have randomly sampled 10\% of users with anorexia and 10\% of control users and calculated the KL-divergence of each individual language model against reference language models for positive and control, and report the average values of each class. We observe that users with anorexia consistently exhibit a lower divergence with respect to positive reference language model than to the control one (0.41 vs.~0.49). Conversely, control users showed a higher similarity with the control reference language model than with the positive one (0.82 vs.~0.90). We have described this proof of concept on Section 5.1.

\subsection{Linguistic Style and Psychometric Attributes}
\label{sub_sec:liwc}



\begin{figure}[!ht]
    \centering
    \subfigure[Reddit]
    {
    \includegraphics[width=\textwidth]{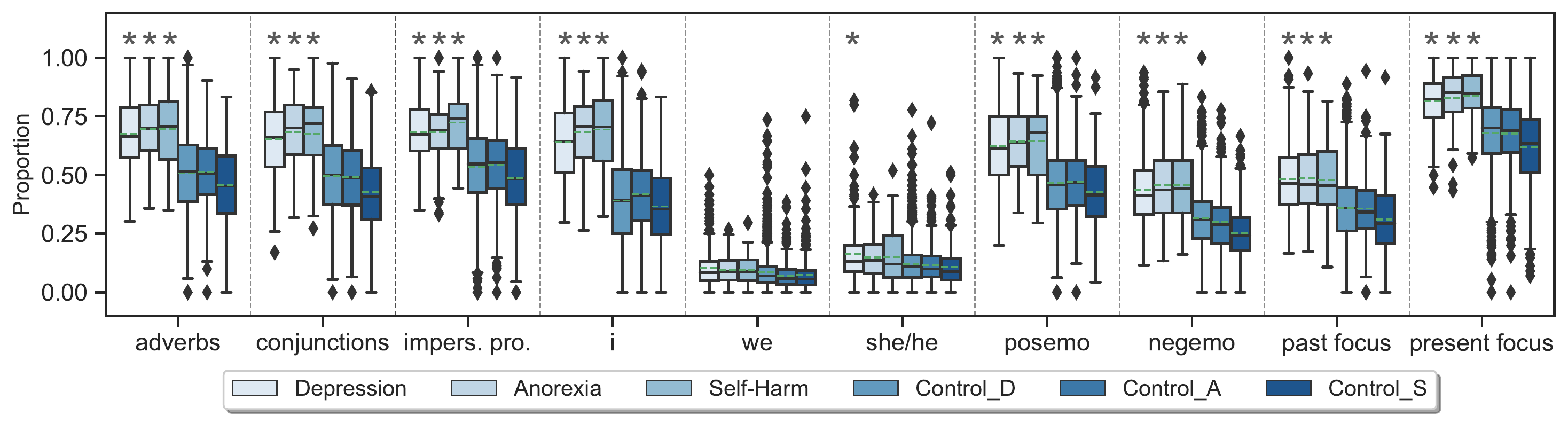}
    \label{fig:liwc_bxplt_selected_1_erisk}
    }
    \subfigure[Reddit]
    {
    \includegraphics[width=\textwidth]{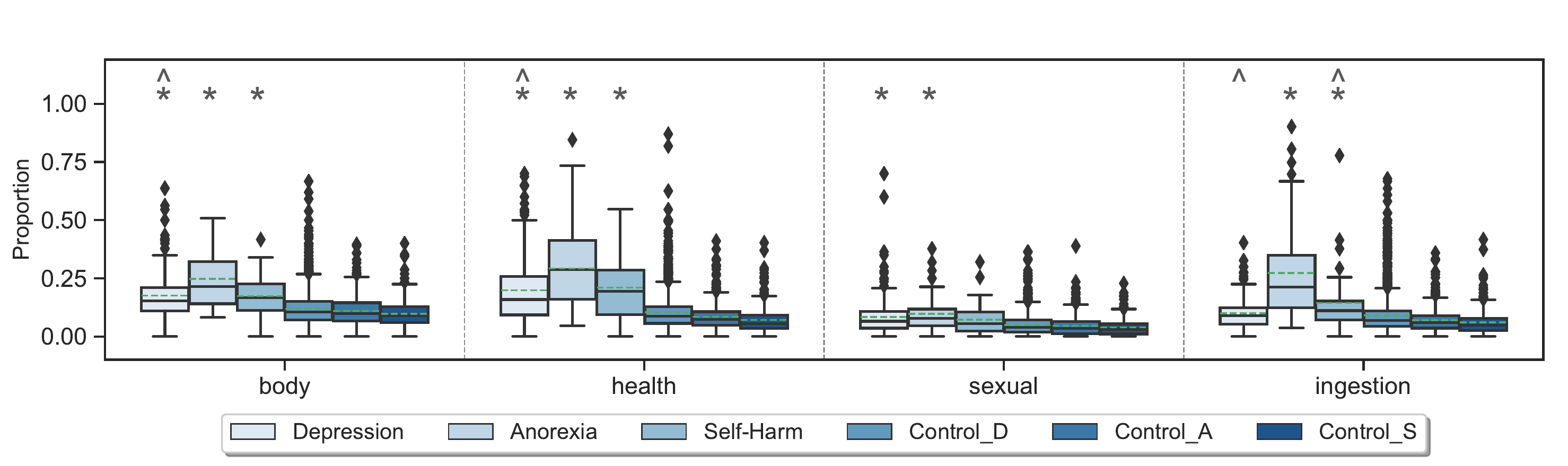}
    \label{fig:liwc_bxplt_selected_2_erisk}
    }
    \subfigure[Twitter]
    {
    \includegraphics[width=\textwidth]{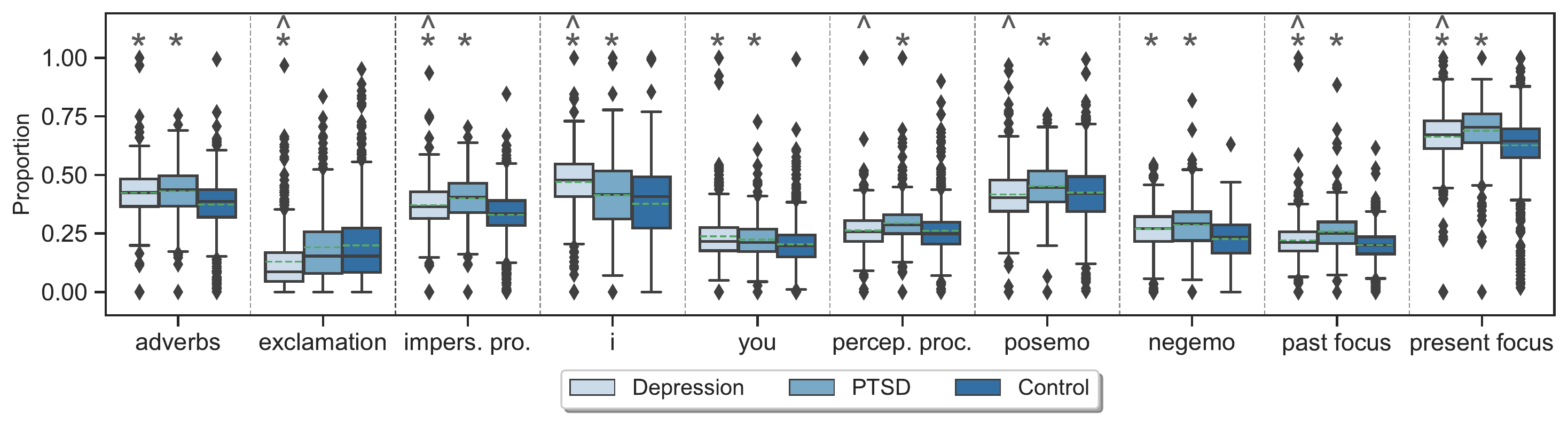}
    \label{fig:liwc_bxplt_selected_1_clpsysch}
    }
    \caption {Box and whiskers plot of the proportion of documents that users have on Reddit and Twitter (y-axis) matching selected LIWC categories. Statistically significant differences between each positive and their respective control groups are denoted by \textbf{*} ($p$-value $< 0.001$). Also, statistically significant differences between Depression/Self-Harm and Anorexia on Reddit and Depression/PTSD on Twitter are denoted by \textbf{\^} ($p$-value $< 0.001$).}
    \label{fig:liwc_bxplt_all_selected_1}
\end{figure}

By comparing a selected set of categories from LIWC and Empath\footnote{Due to space constraints and since trends on Empath support those of observed on LIWC, we do not include the corresponding figures.}, we show that language use has significant differences between positive and control users on Reddit and Twitter. Figure \ref{fig:liwc_bxplt_all_selected_1} depicts the proportion of documents authored by the different users that have at least one word on selected LIWC categories. The bars in the plots are coloured with respect to the class they represent. Selected categories include function words (like pronouns and conjunctions), time orientation (like past focus and present focus), emotionality, drives (like affiliation, achievement and power), as well as cognitive (like insight, tentative and certainty), social (like family and friends) and biological processes (like body and health).

\par Interestingly, we observe that on both social media platforms there is a notorious difference between positive and control users in the use of the personal pronoun ``I''. This replicates previous findings for depression~\citep{Coppersmith:2015a,Choudhury:2013a}.
Moreover, the proportion of messages using words related to positive emotions (named \textit{posemo}) is larger than negative ones (named \textit{negemo}), even for the positive classes, as measured by LIWC on Reddit and Twitter. As literature suggests, a reason for this might be that English words used in everyday language are biased towards positivity~\citep{Kloumann:2012}. Furthermore, this phenomena known as Pollyanna effect~\citep{Dodds:2015} has also been observed in similar studies~\citep{Bathina:2021}. The differences identified on Reddit using LIWC and Empath between positive and control groups are statistically significant when applying a Welch two-sample t-test ($p$-value $< 0.001$). The only exceptions in LIWC are categories \textit{we} and \textit{she/he} as observed on Figure \ref{fig:liwc_bxplt_selected_1_erisk}. On Empath categories affect, nervousness, shame, optimism, love, pain, suffering, violence, death and healing yield significant differences between positive classes and their respective control groups. In contrast, between depression, anorexia and self-harm we were not able to find any significant differences among these categories and those in Figure \ref{fig:liwc_bxplt_selected_1_erisk}. Similarly, in most of the categories analysed with LIWC on Twitter illustrated in Figure \ref{fig:liwc_bxplt_selected_1_clpsysch} statistically significant differences are observed between at least one positive group and its respective control. Using Empath we observe this on categories posemo, negemo, violence, shame, feminine, swearing, communication and death. Moreover, we also see significant differences between individuals suffering from depression and PTSD, including cognitive and social processes and drives for LIWC and violence, feminine, death, communication and swearing for Empath.

\par In Figure \ref{fig:liwc_bxplt_selected_2_erisk} we also showcase LIWC categories relative to \textit{biological process}. This group of categories is comprised of words which are connected to the body and its essential functions. We observe that users with anorexia exhibit a particular use of these words when they are contrasted against depression and self-harm classes. Intuitively, this outcome is expected since anorexia is characterised by an intense fear of gaining weight and a distorted perception of the body image. In general, individuals affected by anorexia are highly likely to retain under control their weight and physical shape, which can pose a strong negative effect on their lives. Such individuals may, therefore, discuss themes about their bodies and its functioning more frequently. In this regard, we found that categories \textit{ingest}, \textit{body} and \textit{health} have statistically significant differences ($p$-value $< 0.001$) between anorexia and depression. In the case of self-harm, only \textit{ingest} category is significantly different from anorexia. Furthermore, LIWC observations are complemented and supported by Empath's categories body, eating, exercise, and ugliness.

\par Finally, we extend LIWC built-in categories with other domain-specific lexicons. In particular, we consider ~\cite{Choudhury:2013a} depression lexicon\footnote{Examples of words included in the lexicon: \textit{insomnia}, \textit{grief}, \textit{suicidal}, \textit{delusions}.} which consists of words closely related to texts authored by people sharing their experience on mental disorders or its symptoms on online settings and also include words associated to names of medications.
We find that, for both Reddit and Twitter, the use of such terms does not significantly differ between users with depression and the rest of the mental disorders studied. This evidences that users suffering from either self-harm, anorexia or PTSD often employ such words as well.
Similarly, we were not able to find any significant differences with the set of absolutist terms\footnote{Examples of absolutist terms: \textit{absolutely}, \textit{constantly}, \textit{definitely}, \textit{never}.} retrieved from the research 
conducted by ~\cite{Al-Mosaiwi:2018}. 

\par Overall, we observe less quantifiable differences between positive and control groups on Twitter than on Reddit. It appears that the use of language on Twitter is less distinguishable between users potentially suffering a mental disorder. This could be related to the limited length or the depth of topics that are discussed on Twitter, as opposed to Reddit. As literature has shown brevity has an impact on the writing style exhibiting distinctive linguistic features; for instance, length constraints disproportionately preserve negative emotions and articles, adverbs, and conjunctions have the highest probability of being omitted~\citep{Gligori:2019}. However, we see that among the two disorders in the data (i.e., depression and PTSD), individuals diagnosed with PTSD are potentially easier to identify by their language use. Also, in general we observe on Twitter similar trends to Reddit, where individuals with depression use a language that resembles that of control users. Even though both PTSD and depression groups are exhibiting similar language use on Twitter, we see that the depression group is even more similar to the control group.

\par As shown, statistically significant differences are observed between positive and control groups (on both social media platforms) as quantified by several psychometric and linguistic style attributes. Using two different though complementary dictionary-based tools, we have shown that this close-vocabulary analysis provides a reliable way to collect quantitative data from user-generated content about the mental state of individuals on social media.


\subsection{Emotional Expression}
\begin{figure}[!ht]
    \centering
    \subfigure[Reddit (eRisk)]
    {
    \includegraphics[width=1\textwidth]{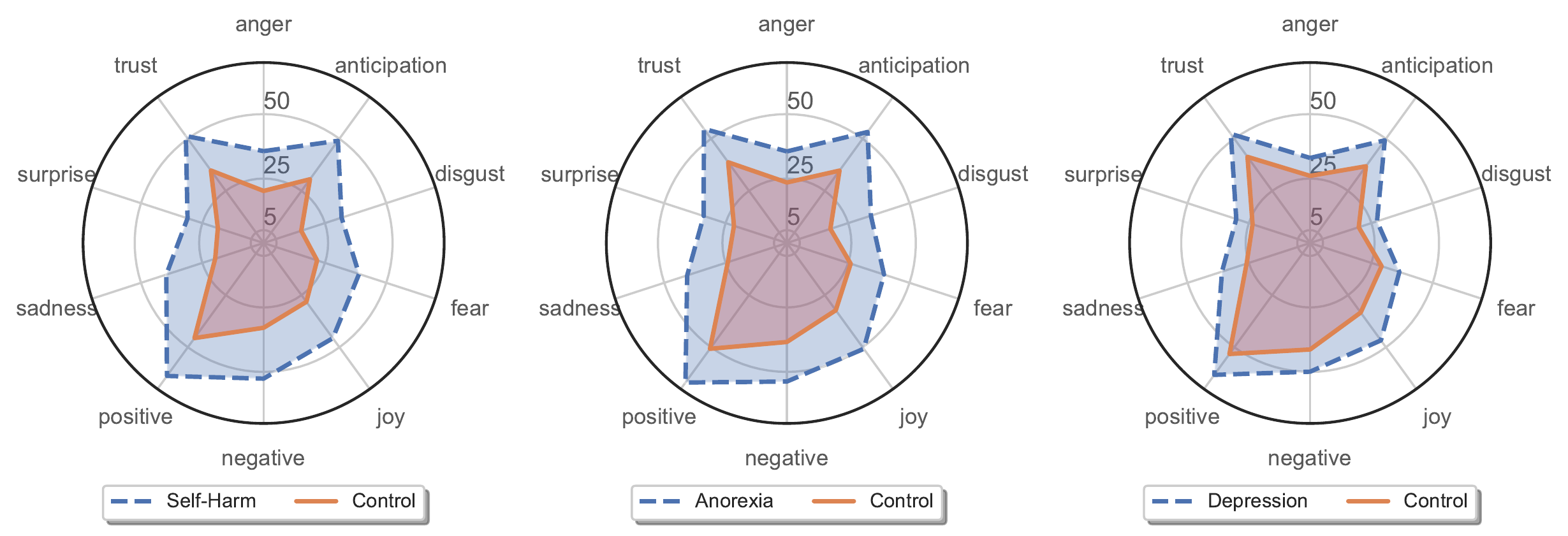}
    \label{fig:nrc_radar_emo_erisk}
    }
    \subfigure[Twitter (CLPsych)]
    {
    \includegraphics[width=1\textwidth]{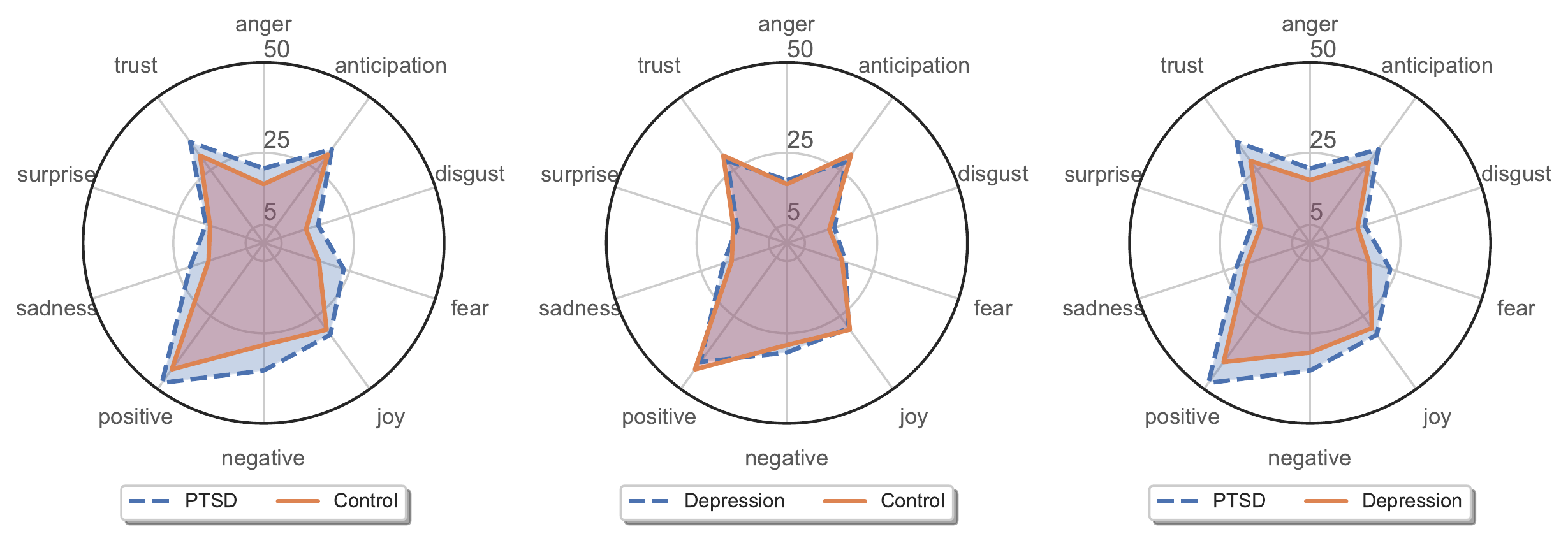}
    \label{fig:nrc_radar_emo_clpsych}
    }
    \caption{Graph depicting the average number of documents containing one or more words associated with a particular emotion, as well as the polarity (positive vs.~negative) for each positive group and its control.}
    \label{fig:nrc_radar_emo}
\end{figure}

Based on the average number of documents that contain at least one word associated to a particular emotion (including the polarity), Figure~\ref{fig:nrc_radar_emo} provides a pairwise comparison between the different positive classes and their respective controls. On Reddit, there are considerable differences between the emotional expressions of the classes under analysis. These results suggest that, on average, affected users tend to share emotions more regularly than control individuals. 
Conversely, depressed and control individuals exhibit an overlapping trend on Twitter, being users affected by PTSD the ones who tend to share slightly more emotions on average. Comparing the social media platforms, we observe that terms related to \emotion{sadness} are less frequent on Twitter. Furthermore, words that relate to the positive sentiment are the most frequent ones among the various emotions on both social media. 
These observations on emotional expression support our earlier findings indicating that there are considerable smaller differences in the use of language between positive and control groups on Twitter when compared to Reddit. Moreover, as it was also previously noted, we see that on Twitter users suffering from PTSD reveal a more distinguishable use of language with respect to controls than depressed individuals.

\par
\begin{figure}[!ht]
    \centering
    \includegraphics[width=1\textwidth,width=0.95\linewidth]{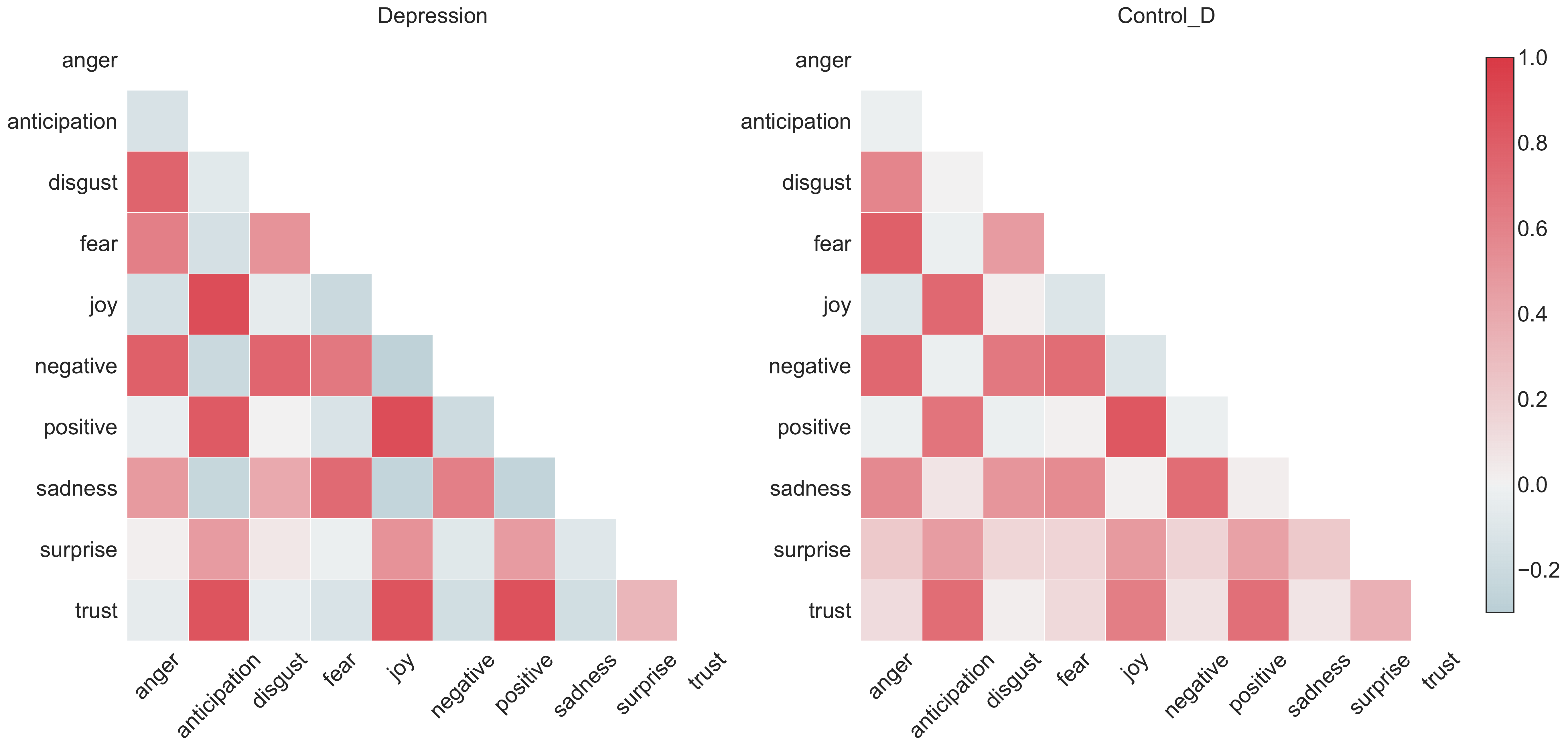}
    \label{fig:nrc_corr_fr_dep_erisk}
    \caption {Heatmap depicting the frequency correlation of the different emotions for positive (left) and control (right) groups on Reddit (best viewed in colour).}
    \label{fig:nrc_corr_fr_erisk}
\end{figure}

\begin{figure}[!ht]
    \centering
    \includegraphics[width=1\textwidth,width=0.95\linewidth]{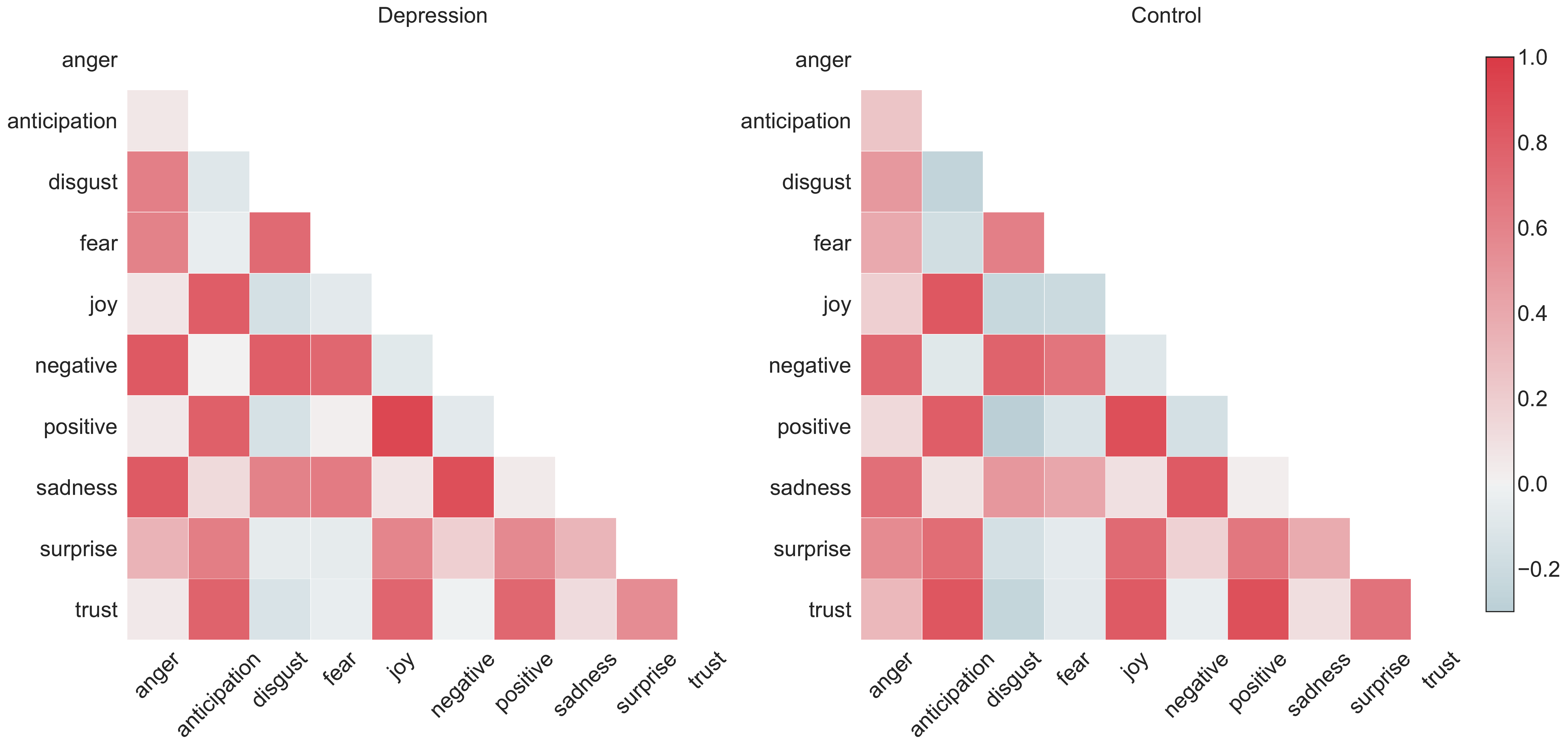}
    \label{fig:nrc_corr_fr_dep_clpsych}
    \caption {Heatmap depicting the frequency correlation of the different emotions for positive (left) and control (right) groups on Twitter (best viewed in colour).}
    \label{fig:nrc_corr_fr_clpsych}
\end{figure}

 Additionally, Figures \ref{fig:nrc_corr_fr_erisk} and \ref{fig:nrc_corr_fr_clpsych} depict a frequency correlation analysis of the various emotions comparing each class against its relevant control. Due to space constraints, we only include depression. However, similar trends are observed from the remaining positive classes. We observe that different classes exhibit different correlations with certain emotions. Interestingly, \emotion{sadness} shows a negative correlation with \emotion{joy} and \emotion{positive polarity} for depression (Figure \ref{fig:nrc_corr_fr_erisk}). However, such correlation is not observed for the respective control group. As another example, we see that for self-harm's respective control class \emotion{surprise} presents a larger positive correlation with \emotion{joy}, \emotion{trust}, as well as with \emotion{positive} and \emotion{negative} polarities when compared to self-harm class. The opposite trend is observed between \emotion{surprise} and \emotion{disgust} as well as between \emotion{fear} and \emotion{disgust}. Interestingly, when we compare the collective correlation of surprise with all other emotions, we see a considerable difference between Control and Depression groups. While the Control group exhibits a positive correlation with all other emotions, the Depression group exhibits more fluctuation. We believe that such observations are worthy and useful to the community. Through studies of such nature, we can better understand how certain emotional patterns arise as a result of emotional expression.

In order to compare both social media platforms we take the particular case of depression. We see in both platforms that \emotion{anger} correlates positively with \emotion{disgust}, \emotion{fear}, \emotion{negative polarity} and \emotion{sadness}. Interestingly, we observe that the positive correlation between \emotion{anger} and \emotion{sadness} is nearly double on Twitter compared to Reddit. In addition, we observe that \emotion{anticipation} correlates positively with \emotion{joy}, \emotion{positive polarity}, \emotion{surprise}, and \emotion{trust} in both social media platforms. We also find notable correlations between \emotion{disgust} and \emotion{fear}, \emotion{negative polarity}, and \emotion{sadness} (larger positive correlation on Twitter). Among other interesting emotions we can point out the positive correlation that \emotion{fear} has with \emotion{negative polarity}, and \emotion{sadness}. 
We also observe more intuitive correlations such as the positive correlation between \emotion{positive polarity} and \emotion{joy}, \emotion{surprise}, and \emotion{trust}. More on the negative sentiment, we see that \emotion{sadness} positively correlates with \emotion{negative polarity} (stronger on Twitter).
Overall, we note that emotion correlations are quite similar on both social media platforms for the various mental disorders. As we observed in other experiments, Reddit users affected by a mental disorder show a more distinctive behaviour from control ones than what we observe on Twitter. However, observing similar correlations implies that even though the bag-of-words analysis of the emotions lead to subtle differences, more sophisticated analyses can reveal more distinctive differences.

\subsection{Social Engagement and Posting Trends}
In Table~\ref{tab:dataset} we present various statistics about the collections studied in this research. Overall, we observe that positive users on Reddit create on average fewer documents, though longer, than control users. The case of users with anorexia is particularly interesting. Their documents are even longer than those written by users who are affected by self-harm or depression. It is good to out-stand that we did not identify meaningful differences among the various control groups. This trend is consistent over various experiments and it is expected as each control group entails a random sample of Reddit posts at different temporal periods. Hence, they constitute a representative sample of Reddit users.

\par In the case of CLPsych collections, we observe very subtle differences between positive and control individuals. The average number of documents generated by each group is nearly the same, being marginally larger for the positive ones. Individuals affected by PTSD write on average slightly longer documents than depressed and control users. Finally, we observe a noticeable difference in the average number of documents per individual between the platforms. Regardless of whether they are potentially affected by mental disorders, Twitter users consistently generate more documents than Reddit users. Furthermore, the difference observed in terms of average document length is clearly a result of Twitter's character limit, which does not exist on Reddit.


\par
\begin{figure}[!ht]
    \centering
    \subfigure[Reddit (eRisk)]
    {
    \includegraphics[width=1\textwidth]{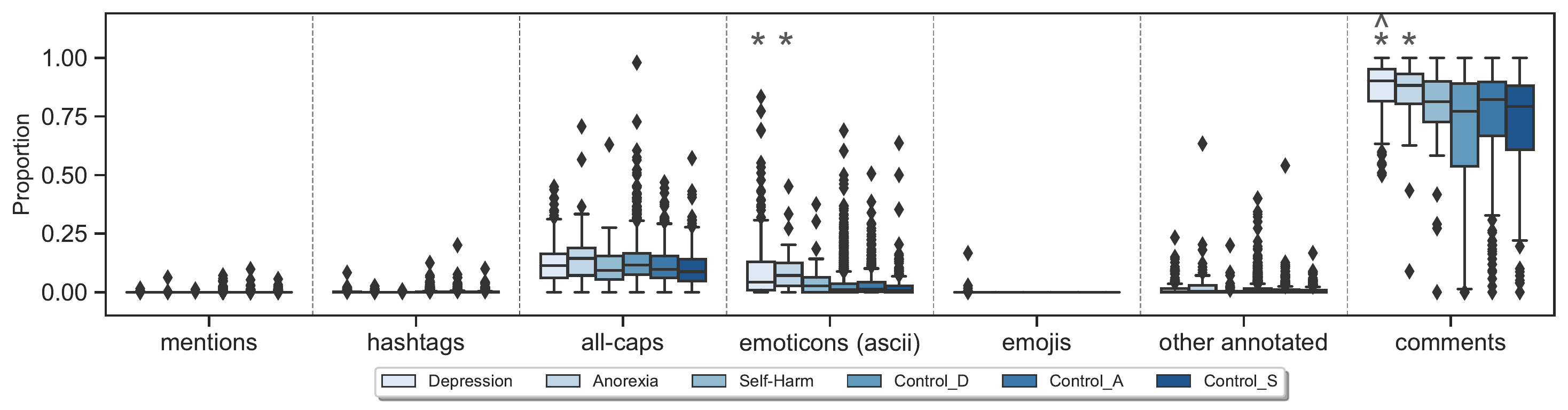}
    \label{fig:behaviour_bxplt_erisk}
    }
    \subfigure[Twitter (CLPsych)]
    {
    \includegraphics[width=1\textwidth]{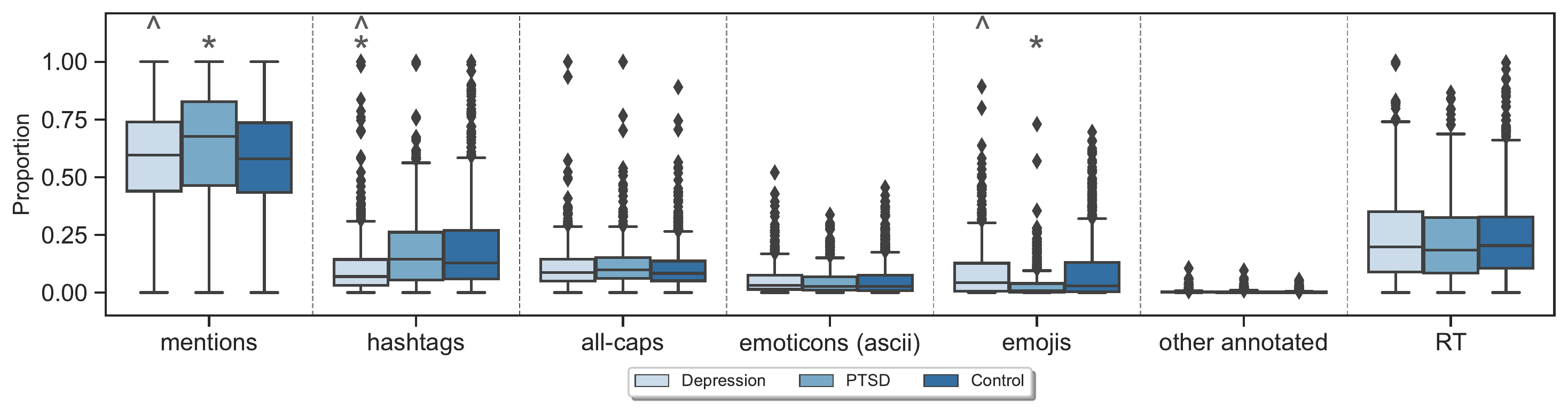}
    \label{fig:behaviour_bxplt_clpsych}
    }
    \caption {Box and whiskers plot of the proportion of documents each user has (y-axis) matching various social engagement and posting trends indicators. Statistically significant differences between each positive and their respective control groups are denoted by \textbf{*} ($p$-value $< 0.001$). Also, statistically significant differences between Depression and Self-Harm (eRisk) and Depression and PTSD (CLPsych) are denoted by \textbf{\^} ($p$-value $< 0.001$).}
    \label{fig:behaviour_bxplt_all}
\end{figure}

 Figure \ref{fig:behaviour_bxplt_all} depicts the proportion of documents that each user has considering various social engagement indicators (i.e., there is at least one occurrence of that element in a document). We considered a variety of elements modelling various aspects of users' interaction and engagement with the social media platform, such as hashtags, mentions, and ASCII emoticons. Bars are coloured according to the positive and control classes they represent.

\par We observe a noteworthy difference between the members of each social media platform. While Reddit users hardly utilise most of the elements studied, their usage is quite common among Twitter members, in particular mentions and hashtags. Interestingly, the majority of the submissions from individuals suffering from depression on Reddit are comments (instead of posts). This suggests that, with respect to control users, most of the time they tend to reply to existing submissions (either another comment or a post) instead of initiating the conversation thread themselves. Hence, in comparison with users who are estimated to be healthy, one could assert that they exhibit a more \textit{passive} behaviour in terms of content creation. On Twitter, hashtags are significantly less often by depressed users. A closer inspection reveals that, on average, users affected by depression exhibit a significantly smaller hashtag ratio ($\approx 1/\%$), meaning that out of ten words one is a hashtag. While for PTSD and control users out of ten words two are hashtags ($\approx 2\%$). Moreover, PTSD users show a significantly larger usage of mentions than depressed and control individuals. This could be interpreted as a way of looking for peer support to overcome their condition. For instance, sharing narratives about the event that triggered their trauma to friends, pals support, or other individuals suffering from PTSD as well. However, there are not significant differences in the ratio of mentions per document among the three groups, being slightly larger for PTSD users ($\approx 7\%$). 
Compared to depressed and control individuals, on average PTSD users tend to include more mentions in their posts, but the frequency of mentions for the three groups in each post is practically equal.


\par
\begin{figure}[!ht]
    \centering
    \subfigure[t][Reddit (eRisk)]
    {
    \includegraphics[width=0.47\textwidth]{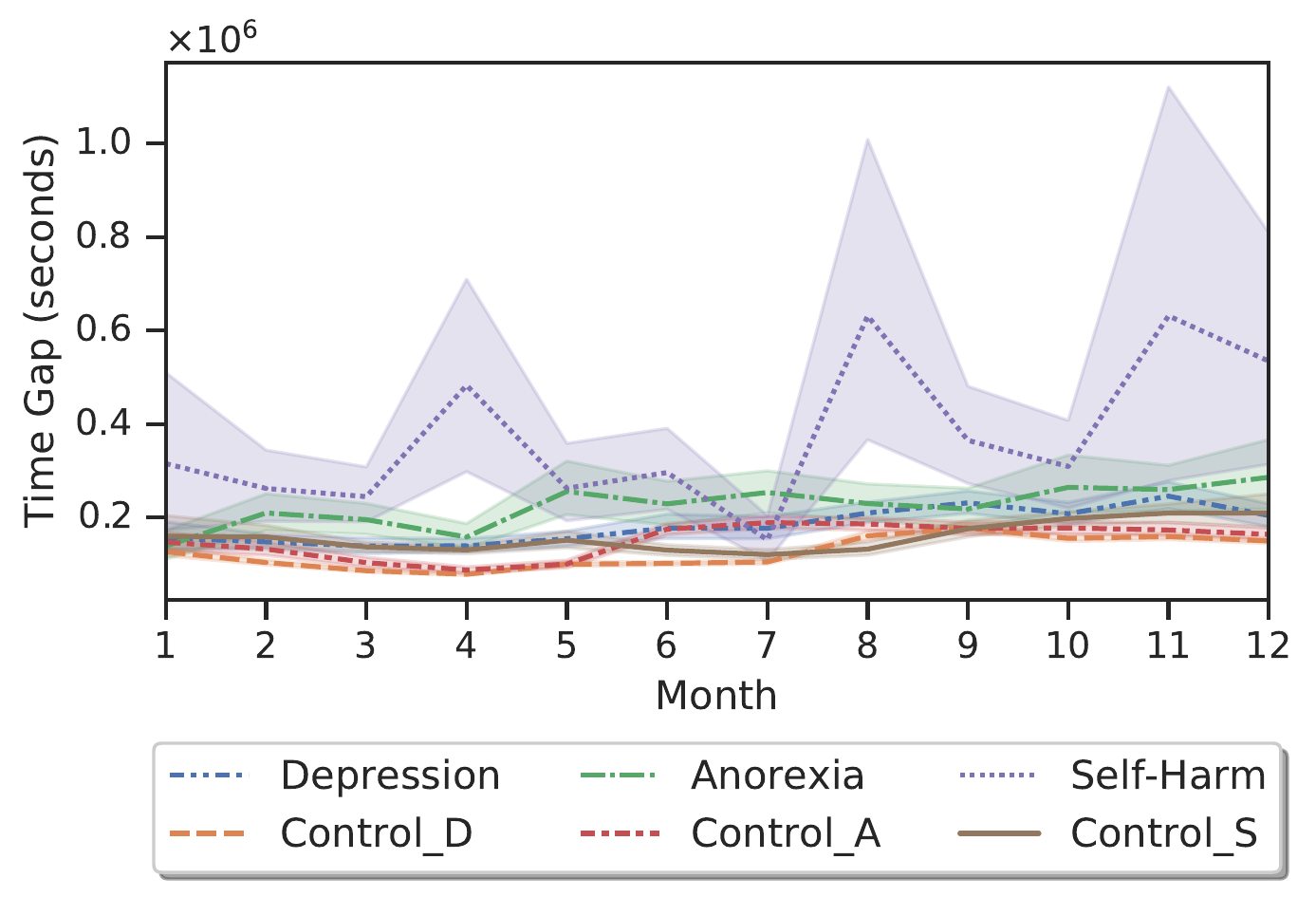}
    \label{fig:behaviour_timegap_erisk}
    }
    \subfigure[t][Twitter (CLPsych)]
    {
    \includegraphics[width=0.47\textwidth]{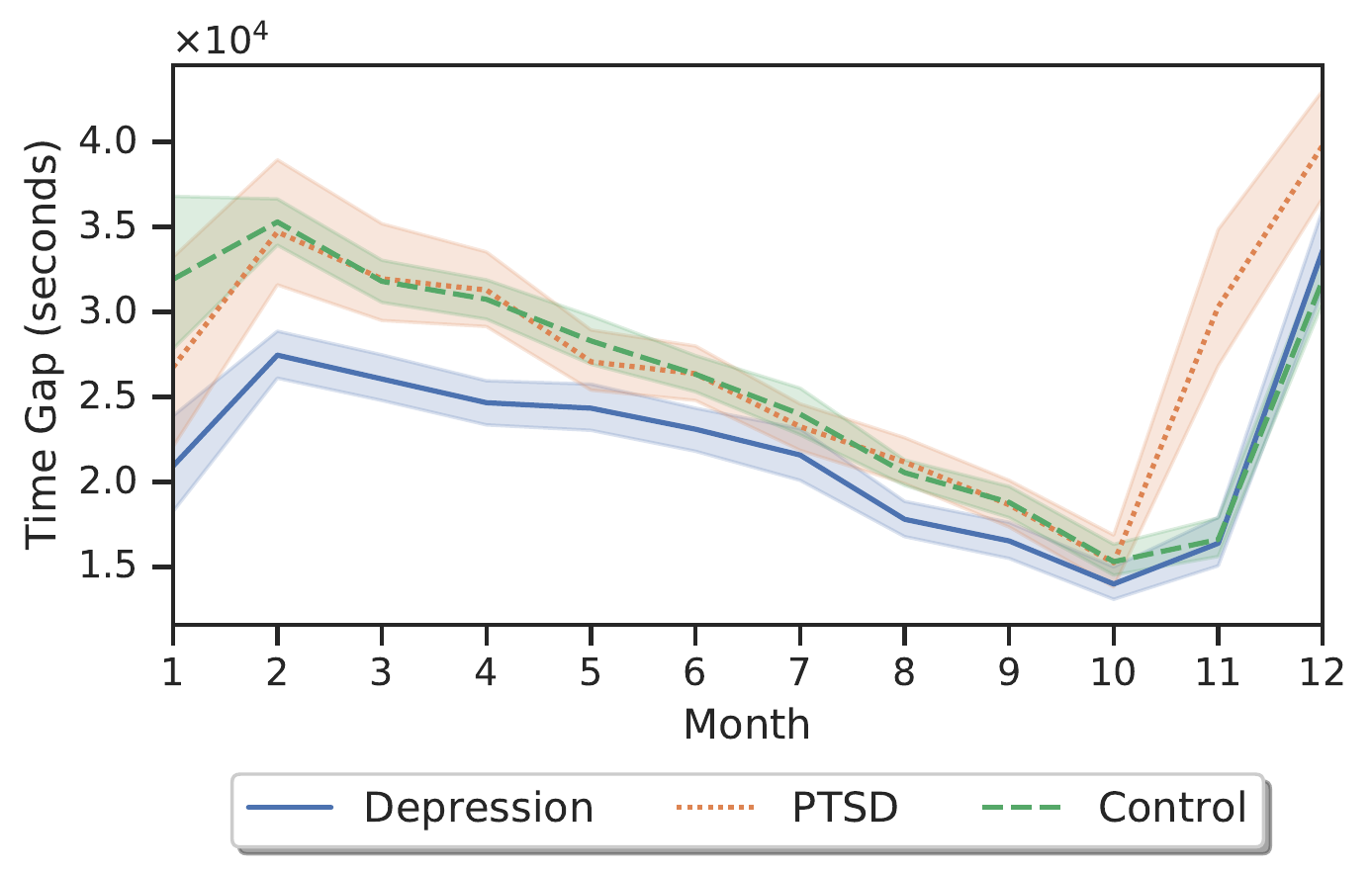}
    \label{fig:behaviour_timegap_clpsych}
    }
    \caption {Time-gap analysis by month.}
    \label{fig:behaviour_timegap_all}
\end{figure}

 An appealing feature of the collections studied in this work is that the average lapse time between users' first post (i.e., oldest interaction) and users' last post (i.e., newest interaction) cover more than a year. In this way, we can study how users' engagement in the social media platform develops in relationship with their mental health condition. Figure~\ref{fig:behaviour_timegap_all} plots the time-gap behaviour of users belonging to different groups in both social media platforms. We plot the average posting time-gap per month for both platforms. In Figure~\ref{fig:behaviour_timegap_erisk}, there is a clear difference in terms of posting regularity of users in different groups on Reddit. In particular, users diagnosed with self-harm exhibit the highest variance on a monthly basis. We see that while the time-gap in some months (e.g., April, August, and November) is very high, the users post more frequently in other months. Moreover, we observe from the highest standard deviation in the behaviour of users that individuals who suffer from self-harm exhibit the least posting regularity. This could be due to the so-called \textit{self-harm cycle}\footnote{Inflicting some sort of pain can bring a momentary sense of calm to the negative feelings experienced by the individual. Although, this is often temporary and can lead to feelings of guilt and shame which can stimulate again the original emotions leading to further self-harm.}. On Reddit data, we also see that users who suffer from Anorexia exhibit a larger time-gap between their posts. Moreover, the monthly posting variance is higher than the control groups. In general, we observe that the control groups show a relatively stable posting behaviour. We cannot see any noticeable differences between the three control groups on Reddit. Similarly to other measures, we see less distinction between the depression group and the control groups on Reddit. However, as we see in Figure~\ref{fig:behaviour_timegap_clpsych}, users on Twitter who suffer from depression, seem to post more frequently. We observe that the control group posting trend is more similar to the PTSD group. Interestingly, towards the end of the year (i.e., November and December), the control group users behave more like the depression group users. Finally, we observe in the plot that users belonging to the PTSD group exhibit a higher variance (i.e., the higher standard deviation in the graph), indicating less stability compared to the control group, even though the average time-gap is nearly the same.

\par As observed on the time-gap analysis conducted certain behavioural patterns related to specific periods or events that take place during the year, e.g., seasons change or Christmas and New year's eve emerged. Moreover, we also have an intuition that there could be some individuals who experience seasonal affective disorder (SAD) throughout the year. Many people go through short periods of time where they feel sad or not like their usual selves. Sometimes, these mood changes begin and end when the seasons change. This can be more acute for individuals who are already suffering from depression or other mental disorders. However, as posts are not geo-located, we cannot directly identify such cases yet some patterns still manifest. The fact that there are certain periods of the year in which different trends emerge, such as a higher posting volume at specific intervals, suggests that some event/s negatively affected users during those periods. We believe that this information could be useful for practitioners and medical agencies to launch prevention campaigns and encourage individuals to visit a professional to talk or discuss about their mood, feelings and mental condition during that time. This could foster interventions at early stages and avoid any further consequences. In addition, it could prevent the emergence of new cases. In fact, as observed on the plot, control users on Twitter exhibit a similar behaviour to depressed users towards the end of the year. This could be the result of family events or even the lack of them and hence it could negatively affect individuals mental health. On some users, this could trigger over time other episodes. Finally, if automatic systems are used to identify potential cases, their thresholds and decision rules could become more \textit{sensitive} during these periods and, thus, raise a larger volume of alarms.

\section{Implications, Conclusions, and Future Work}
\label{sec:conclusions_future_work}


\par Continuously generated content on social media yields new opportunities for innovative analysis on mental health concerns. Social media data is produced in a natural way by users' own willingness and therefore, has become a complement of great value for traditional assessment instruments (like self-assessment inventories) which are employed to establish the likeliness of occurrence of mental health concerns. 
Unlike most of the work in the area that aims at predicting the mental state and solving such a complex problem without the help of a practitioner, we follow our previous work~\citep{Rissola:2020}, focusing on providing useful means of diagnosing mental health patients by the practitioners. In particular, we proposed novel ways of consuming and visualising the vast amount of textual and behavioural data to extract indicative cues to the practitioners. In this work, we extended our initial findings~\citep{Rissola:2020} on the Reddit platform to Twitter and studied how our analysis can be generalised to other social media platforms. Moreover, we focused not only on the textual data, but also behavioural data on both social media platforms.

\par Outcomes from our thorough analysis revealed significant differences in writing and behavioural patterns on social media platforms between individuals with mental disorders and control users. The visualisation and study of certain probabilistic attributes enabled us to detect trends in affected users' writing style, emotional expression, and online behaviour on social media. In particular, we observed and researched the vocabulary, psychometric and emotional indicators, and activity of different users on two social media platforms of diverse characteristics. While analysing and visualising these attributes, we found interesting differences that could be helpful for the development of diagnosis systems, and, specially, to provide health professionals with a more comprehensive way to assess individuals' mental state. However, we were not able to find any significant and distinctive indicators among the different mental disorders to differentiate from one another. 
We obtained evidence suggesting that the use of language in micro-blogging platforms, like Twitter, in which users are subject to constraints influencing their writing style, is less distinguishable between users suffering a mental disorder than other less restrictive platforms, like Reddit.
As a result, we come to the conclusion that social media posts could be analysed to automatically detect individuals who hold a higher probability of being diagnosed with a mental disorder. Yet, specifying the disorder entails higher complexity and that is the reason why the judgement of an expert may be required. 

\par Implications resulting from our work serve as a starting point for future research dedicated to understanding how mental health concerns are manifested on social media. The various analysis techniques presented in this work and the information they provide could be integrated with more conventional direct and indirect assessment instruments used by mental health practitioners to create a more comprehensive cross-sectional characterisation of the persons' mental condition. As shown throughout the article, when the mental disorder of users develop the way they write and express detaches from the \textit{standard} behaviour. Once potentially affected users are identified, a report could be generated to estimate how much their writing differs from healthy individuals in an attempt to characterise their condition. For instance, one dimension to consider in such report could be the close-vocabulary analysis presented in Section \ref{sub_sec:liwc}. If the users consistently show that they have increased the usage of words in the selected LIWC and Empath categories, this could provide stronger evidences that they are affected by a mental disorder. Similar scenarios could be described considering the rest of the dimensions studied in this work.


We plan to investigate various directions as future work. Firstly, we aim to expand our study to multiple datasets collected from different social media platforms to provide more insights into the behaviour of users with various mental issues on different social media platforms. Moreover, this would allow us to conduct a thorough analysis on the impact of different datasets on the results that we obtain. To do so, we aim to characterise each dataset with a number of features such as having length limitations and see how such limitations or differences across social media platforms would affect users' behaviour. Secondly, while in this work we have mainly studied the lexical nature of social media posts, in the future we plan to extend to semantic models such as word2vec~\citep{Mikolov:2013} and contextual embeddings\citep{Devlin:2018}, as well as incorporating multimodal information in our analysis such as multimedia context (e.g.~images and videos)~\citep{Gkoumas:2021}.
Finally, we plan to study the effectiveness of the analyses we performed in this work on machine learning approaches by using them as features to predict the individuals with mental health issues.
\section*{Acknowledgements}
This work was supported in part by the Swiss Government Excellence Scholarships and the Hasler Foundation, 
and in part by the NWO Innovational Research Incentives Scheme Vidi (016.Vidi.189.039).

\bibliography{main}

\end{document}